\documentclass{york-thesis-modified}

\usepackage[tracking]{microtype}
\usepackage[T1]{fontenc}  
\usepackage[utf8]{inputenc} 
\usepackage{booktabs}
\usepackage{amsfonts}
\usepackage{hyperref}
\usepackage{algorithm}
\usepackage{algorithmic}
\usepackage[pdftex]{graphicx} 
\usepackage{amsmath, amssymb}
\usepackage{yfonts}
\usepackage{commath}
\usepackage{longtable}

\usepackage[
backend=biber,%bibtex
style=ieee,
sorting= none %ynt
]{biblatex}
\newcommand{\cG}{\mathcal{G}}
\newcommand{\cD}{\mathcal{D}}
\newcommand{\bz}{\mathbf{z}}
\newcommand{\bx}{\mathbf{x}}
\newcommand{\fo}{\mathcal{F}}
\newcommand{\cL}{\mathcal{L}}
 \setboolean{masters}{true}      % set true for a Master's thesis; % false for a PhD dissertation
 \setboolean{hasfigures}{true}  % set true if you have figures
 \setboolean{hastables}{true}   % set true of you have tables

 \title{Explicit Use of Fourier Spectrum in Generative Adversarial Networks}       
 \author{Soroush Sheikh Gargar}   % this is your name exactly as you want it to appear everywhere in your work.
 \department{Electrical Engineering \& Computer Science}   % this is the name of the programme in which you are attempting your degree
 \masterof{Applied Science}        % this is the type of Master's degree you are attempting. If the boolean masters is false, this has no effect.
 \degreename{Masters}       % put the actual name of the degree you are attempting {Doctor or Masters}
 \date{September 2021}     % this is the month and year of defence

 \committeememberslist{
   \begin{enumerate}
     \item Hui Jiang
     \item Mokhtar Aboelaze
   \end{enumerate}}

\abstractfile{abstract.tex}
\acknowledgementsfile{acknowledgements.tex}
\abbreviationsfile{abbreviations.tex}

\begin{document}
    \makefrontmatter
    % \maketitle
    % \input{abstract}
    \chapter{Introduction}
\label{chap:introduction}
Generative adversarial networks (GANs) proposed by \cite{GAN_original} have had a significant effect on the computer vision literature. They have been used in a wide range of applications, such as generating realistic images \cite{DCGAN, styleGAN, styleGAN2}, image-to-image mappings \cite{pix2pix}, text-to-image translation \cite{text2Image0, text2Image1, text2Image2, text2Image3}, transfer learning \cite{few}, photograph editing \cite{edit0,edit1,edit2,edit3}, 3D object generation \cite{3dgan0, 3dgan1} and many more that are still coming up. Image synthesis state-of-the-art works successfully generated photo-realistic images almost indistinguishable for human eyes. This rapid progress gave birth to the applications such as deepfake \cite{deepfakes} in which the central idea is to fool the human eyes into believing the machine-generated images as real ones. Naturally, it raised security concerns on the problem challenging photo-forensics experts on the subject \cite{photoforensic}. On the flip side, recent works \cite{easytospot} have shown that having a "large" amount of data, one can detect fake images with relative ease using simple classifiers powered by neural networks. \cite{Zhang_2019} attributes the machine's power for classifying to the dissimilarities between the authentic and artificial images in their spectrum. It further discusses the roots of dissimilarities in the high-frequency residuals of the image synthesis in convolutional neural networks. It also discusses that the checkerboard effect, one of the most disputed phenomenons in the GANs, is one of the consequences of the frequency discrepancies.\\
The Frequency domain is a well-established tool to process and analyze the images in the signal processing literature. It represents images without losing any information due to the bijective property of the FFT. This representation gives us the advantage when dealing with the rate of change within the neighborhood of the pixels, intuitively making it related to the problem of the checkerboard patterns and any other harsh changes regarding the details of an image.\\
%%% revision
The main objective of the current research is to find a deep learning based method to reduce the existing gap between the frequency spectrum of the real images and the ones produced by the generative adversarial networks. We also ask the question of how we can find a way to incorporate frequency information of the images in the deep learning driven computer vision tasks. To be precise, to build a deep learning layer for processing frequency information along with spatial information.\\
%%%%
Imitating human responses and behaviors has always been a topic of utmost interest for artificial intelligence researchers. Spatial frequency theory \cite{bio_theory} is an acclaimed construct showing the weight of the spatial frequencies in the brain's neural response. Recent research \cite{cortex} also reveals that the mechanism of the human primary visual cortex (V1) is highly related to the second-order spatial frequencies received from our visual sensory inputs. These spatial frequencies stand out in the spectrum representation of an image. Useful filters such as the Gabor filter applied on the spectrum of the images are also considered to be similar to the actions of cells in the mammalian visual cortex for texture discrimination \cite{gabor1, gabor2, gabor3}. Knowing how vital the frequency domain is in the ultimate baseline of the AI, meaning humans, motivates us even more to use models consist of frequency-related elements.\\
To get an image realistic to the finest detail, generative models need to learn the distribution of the images accurately enough to imitate the spectrum of the generated images, which is not the case for the current approaches. Identifying this problem, in this work, we are trying to propose a new add-on to the current architectures of the CNN-based GANs in order to incorporate the frequency representation with more emphasis in the training process. \\
The rest of this work is organized as follows: In chapter \ref{chap:backgrounds} the necessary information for understanding the ideas are provided. Chapter \ref{chap:literatureReview} introduces the current major advances in the literature of generative adversarial networks and frequency-driven efforts in the area of neural networks. Next, in chapter \ref{chap:GeometricDeepLearning} we discuss the idea of the geometric deep learning, which is the main building block for our contribution. Chapter \ref{chap:ProblemDef_Methodology} formalizes the problem at hand and gives theoretical methodologies for our contributions. Chapter \ref{chap:eval} assess the performance of the idea in spatial and frequency domain as well as a stability analysis of the implementations. Finally, chapter \ref{chap:conclusion} summarizes the contributions of the paper and discusses the future steps for this research.
    \chapter{Backgrounds}
\label{chap:backgrounds}
In this section, we first present the concept of neural networks, and with a simple example of MLP, we explain the procedure behind their ability to learn. Then we discuss Convolutional Neural Networks as the most successful neural network structure in vision tasks. Then we move to Generative Adversarial Networks, which are the main neural network architecture examined in the current work. We explain their structure, the basic mathematical idea behind them and intuitively explain how they are supposed to work. In section \ref{sec:eval} we talk about the methods for GAN's quality evaluation in the spatial domain we used in our work. For evaluation in the frequency domain, we present three metrics later in \ref{sec:FreqEval}. Section \ref{sec:Freq} gives a brief description of the frequency domain and its importance, plus some of its applications in natural language processing and computer vision. It also contains the fundamental mathematics of discrete Fourier transform in 1D and 2D.

\section{Neural Networks}
\label{sec:NN}
\begin{figure}[ht]
\caption{ (a) building blocks of the MLP with input vector $\bx$, weights $w$ and activation function $\sigma$, (b) an MLP structure with a two-neuron output layer and two hidden layers}
\centering
\includegraphics[width=0.8\textwidth]{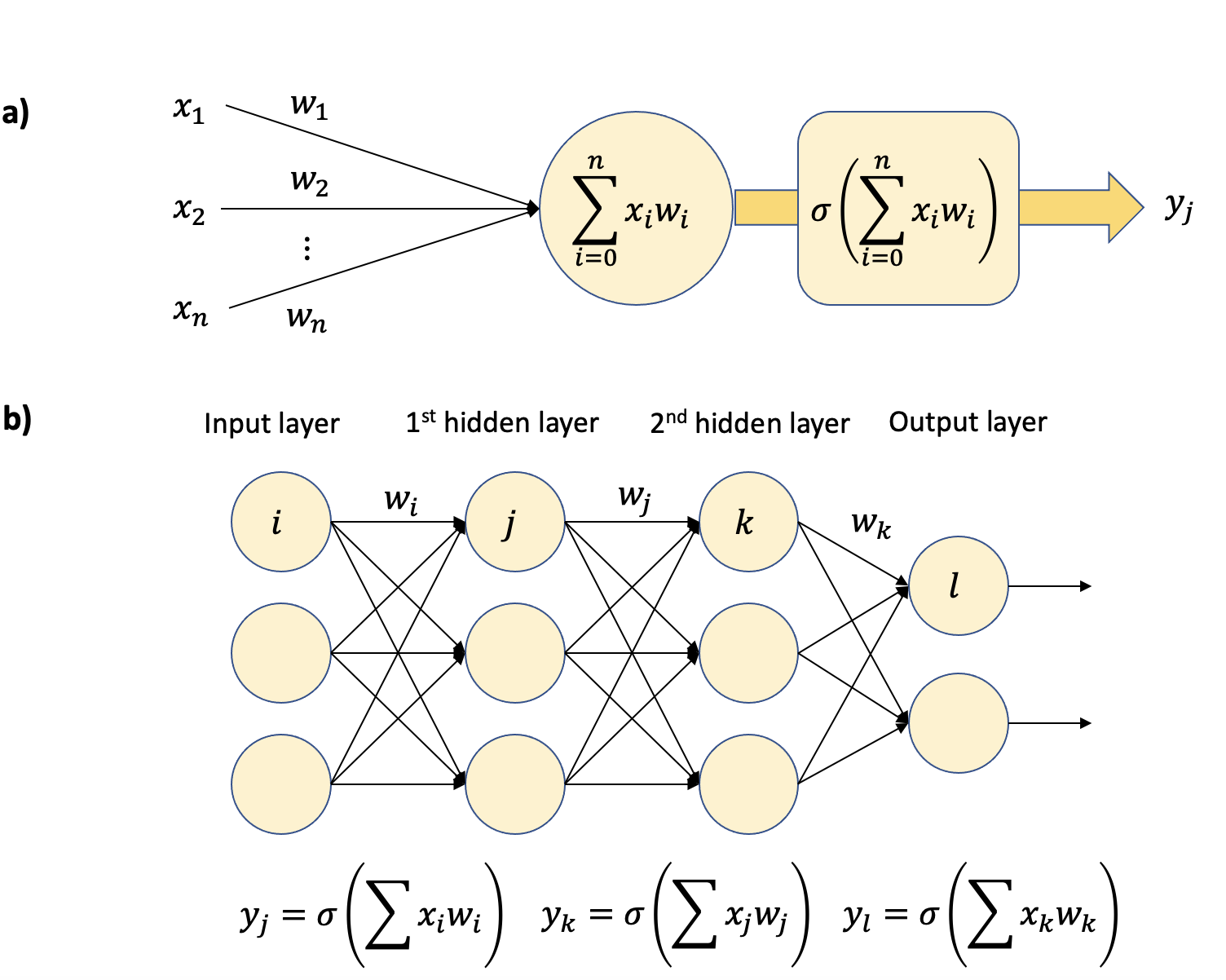}
\label{fig:NN}
\end{figure}

Neural Networks, also known as Artificial Neural Networks (ANNs), are a set of algorithms in machine learning. The recent deep learning revolution in computer science owes its existence to these frameworks of algorithms. They got their name on their design to mimic the structure of human neurons in the brain.  The most simple structure of neural networks, known as Multi-Layer Perceptron (MLP), consists of a layer of nodes (neurons) as the input layer, one or more hidden layers, and an output layer connected with a designed architecture in a layer-wise manner. Values of the nodes in each layer can be seen as a vector and determined by three elements. The first element is the vector of the value of the previous layer. Second, a weight function linearly transforms the previous layer's outputs, represented as a matrix. Finally, a non-linear activation function adding non-linearity to the structure. Figure \ref{fig:NN} shows the architecture a simple MLP with two hidden layers, and activation function $\sigma$.\\
NNs function upon training over a large amount of data, changing their wight matrices gradually with a process called backpropagation to improve their accuracy. The popularity of the NNs is owing to their quick performance and superb accuracy over data-driven tasks such as prediction, classification, and regression.

\begin{figure}[ht]
\caption{ The convolution step with linear generators. (image credit for \cite{geometric})}
\centering
\includegraphics[width=0.8\textwidth]{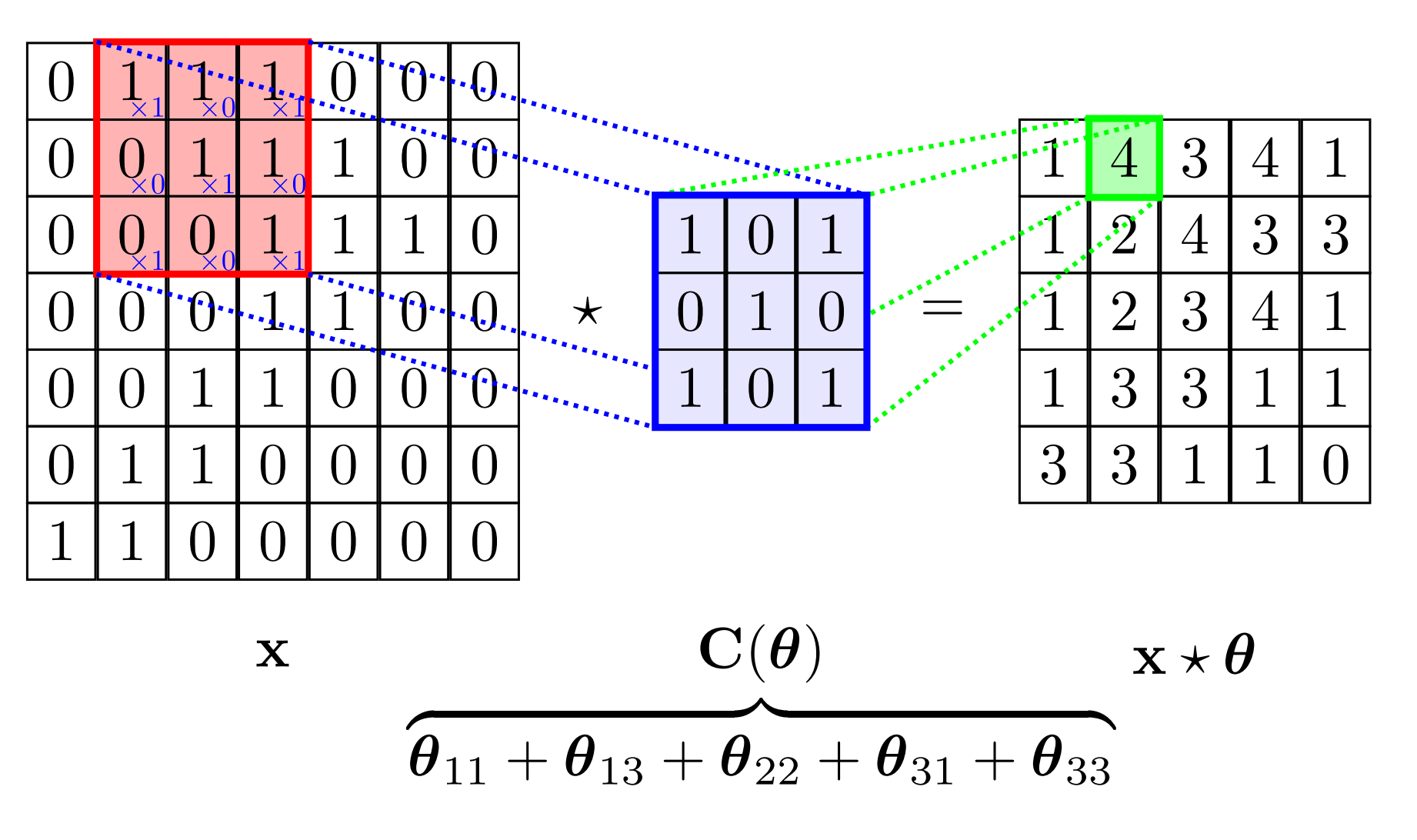}
\label{fig:cnn}
\end{figure}

\paragraph{Convolutional Neural Networks (CNNs)} CNNs are a subset of deep learning algorithms introduced in \cite{cnn}. They are best for learning grid-like data such as Images and time series. They have achieved remarkable accuracy and became an inseparable part of image-related and video-related tasks such as object segmentation \cite{segment0, segment1, segment2}, image classification \cite{classification0, classification1, classification2}, object detection \cite{detection0, detection1, detection2} and image reconstruction \cite{reconstruction0, reconstruction1, reconstruction2}. Moreover, they have shown capabilities in natural language processing such as Sentiment Analysis and Topic Categorization tasks \cite{cnnnlp0, cnnnlp1}, semantic clustering \cite{cnnnlp2} and relation extraction \cite{cnnnlp3, cnnnlp4, cnnnlp5}\\
In a naive two-dimensional convolutional layer, the input should be in a grid form $\Omega = [H] \times [W]$ with vector $\mathbf{u}= (u_1, u_2)$ define over it. An image can be shown as signal in this space $\bx \in \mathcal{X}(\Omega, \mathbb{R})$. A convolutional filter (whose weights are subject to learning) slides over the image. We can write the convolution step with filter size $H^f \times W^f$ using the linear combination of generators $\boldsymbol{\theta}_{1,1}, ... \boldsymbol{\theta}_{H^f , W^f}$. In here lets use the unit impulse function $\boldsymbol{\theta}_{v,w}(u_1,u_2)=\delta(u_1-v, u_2-w)$.
\begin{equation*}
    \mathbf{F}(\bx) = \sum^{H^f}_{v=1} \sum^{W^f}_{u=1} \alpha_{vw} \mathbf{C}(\boldsymbol{\theta}_{vw})\bx
\end{equation*}
$\boldsymbol{\theta}_{vw}$ and $\bx$ are $2$D-matrices flattened into a vector here. Figure \ref{fig:cnn} shows the operation $\mathbf{C}(\boldsymbol{\theta}_{vw})\bx$.
In coordinates, it also corresponds to the standard notation of $2$D-convolution
\begin{equation*}
    \mathbf{F}(\bx)_{uv} = \sum^{H^f}_{a=1} \sum^{W^f}_{b=1} \alpha_{ab} x_{u+a, v+b}
\end{equation*}
When we have more than one channel, e.g., RGB photos or in the later layers when we have several feature maps, a convolutional tensor is used instead of a convolutional filter. And, in coordinates, we have:
\begin{equation*}
    \mathbf{F}(\bx)_{uvj} = \sum^{H^f}_{a=1} \sum^{W^f}_{b=1} \sum^M_{c=1} \alpha_{jabc} x_{u+a, v+b,c}, \quad j \in [N]
\end{equation*}
where M and N are respectively the number of input and output channels.\\
After the convolutional step, the output, known as the feature map, is subjected to a pooling operation which down-samples the feature map to reduce the redundancy and add robustness against over-fitting. Convolutional layers can be cascaded together to form a deep CNN, and they can also be a component of a neural network with linear layers haven their outputs flatten to a vector.\\
Deep-learning-powered computer vision has been thriving in virtue of the abilities CNNs unlock over the normal perceptron. The foremost power in CNNs is their ability to extract features such as patterns, texture, object edges, etc., automatically with convolutional kernels. Some other benefits of using CNNs are as follow: 1) avoiding parameter growth with increasing the number of inputs (let us say pixels for the case of images); this is on account of two facts, the weight sharing over layers, and having local connection which means each node is connected only to some local nodes in the previous layer not all of them. 2) Robustness against translation, in other words, when the input image is slightly shifted, CNN does not treat it as a completely different input. This characteristic, known as shift-equivariency, is explained in detail in chapter \ref{chap:GeometricDeepLearning} 
\section{Generative Adversarial Networks}
\label{sec:GAN}
\paragraph{Generative models}
In general, machine learning algorithms can be categorized into supervised and unsupervised learning. In the former, a labeled dataset is necessary. The goal is to predict an accurate function, mapping the features of the dataset to the target labels. Some examples of it are classification and regression. While in unsupervised learning, there is no labeled dataset; all the data are in the same category. Generative modeling is an unsupervised task in machine learning aiming to learn the data distribution well enough so that the model can produce new data points with all the attributes of the input data. Intuitively, designing a generative model requires a deep understanding of the semantics of the data and the field. However, Generative Adversarial Networks (GANs) are a subcategory of generative models capturing the significant features of the dataset automatically.
\paragraph{Generative Adversarial Networks}Goodfellow et al. \cite{GAN_original} introduced the concept of Generative Adversarial Networks as an algorithmic architecture that uses two neural networks competing against each other. The goal is to generate new data points imitating the distribution of the original training dataset. The skeleton consists of two separate neural networks. A generator $\cG$ that tries to capture training data distribution. Plus, a discriminator $\cD$ whose goal is to predict the probability that its input is a generated fake sample or an original data point. The two networks will train simultaneously to satisfy the following two-player minimax game:
\begin{equation}
    \min_\cG \max_\cD V(\cD,\cG) = \mathbb{E}_{\bx\sim P_{\bx}(\bx)}[\log \cD(\bx)]+ \mathbb{E}_{\bz\sim P_{\bz}(\bz)} [\log (1-\cD(\cG(\bz)))].
    \label{eq:gan}
\end{equation}
$P_\bz(\bz)$ is a defined prior over input noise variable $\bz$, $\cG(\bz;\theta_g)$ is a mapping from noise latent space to the data space implemented with a neural network with parameters $\theta_g$. $\cD(\bz;\theta_d)$ is another mapping from data space to a single scalar with parameters $\theta_d$. $\cD(\bx)$ shows the probability that $\bx$ came from data distribution rather than generator.
The first term of the right-hand side of the equation (\ref{eq:gan}) is an expectation representing the quality of the model performance on detecting the authentic images. While the second term representing the quality of the model on generated samples. Together they build up the loss function $V(\cD,\cG)$, where the discriminator $\cD$ tries to maximize and generator $\cG$ tries to minimize. Training them simultaneously with precise hyperparameter tuning results in both networks getting better at their works in detecting or generating data points. At the final step, the stand-alone generator module will produce indistinguishable data points.
\section{Methods for Measuring GAN's Quality}
\label{sec:eval}
To measure the performance of the neural network models on supervised tasks, researchers use various metrics. For the classification, we have precision, recall, accuracy, F1-score, etc.; for regression tasks, we have mean square error, mean absolute deviation, etc. However, all of them need a labeled dataset out of hand for any unsupervised task, including GANs. Defining a metric for unsupervised tasks is more difficult. Clustering tasks, for example, are usually evaluated using the Silhouette Coefficient. It measures how similar an item in the cluster is to the other items in the same cluster. GANs case is different and even trickier; there is no label in hand, and there is no ground truth other than some samples from the real dataset. The loss for each neural network is not in any way an indicator of how successful the model is in generating a distribution close to the original sampled distribution. They only show how good is each module (Generator or Discriminator) in fooling the other one. Finding an excellent metric to evaluate the GANs performance has been a topic of controversy for a while now \cite{evaluation}. Nonetheless, for quantitative measurement of their success, there are two widely adopted metrics, the Inception Score (IS) \cite{inception} and Fréchet Inception Distance (FID) \cite{fid}.
\paragraph{Inception Score (IS)} IS takes two main high-level concepts into account:
\begin{itemize}
    \item each image decidedly belong to a class (image of a cat belongs to the cat class, and there is no confusion about putting it in another class)
    \item The images are diverse (the model can generate images of a wide variety of classes)
\end{itemize}
To mathematically talk about the concepts as mentioned earlier, using the notion of entropy is useful. The higher the entropy of the data, the less predictable its behaviors. Accordingly, if we get random variable $x$ as the given images and $y$ as the image class, to satisfy the first concept, we want the conditional probability $p(y|x)$ to have low entropy. In other words, predicting the class, given an image, should be a highly predictable task. For the second concept, we want the marginal probability $p(y)$ to have a high entropy; namely, we want the model to give us different classes as uniform as possible. Bringing together the two concepts, we have the final form of inception score using KL-divergence as the following:
\begin{equation}
    \mathbf{IS}(\cG) = \exp(\mathbb{E}_{\bx\sim P_\cG}D_{KL}(p(y|\bx)||p(y))
\end{equation}
For the computation of posterior distributions, the classifier, Inception Net, \cite{inceptionNet} is being used over the data (hence the name Inception Score). Then, the marginal probabilities are calculated by the following integral:
\begin{equation*}
    \int_\bz p(y|\bx = \cG(\bz))d\bz
\end{equation*}

\paragraph{Fréchet Inception Distance (FID)} FID introduced by \cite{fid} embeds a set of generated samples and also actual samples into a feature space. It uses an intermediate layer of Inception Net (usually after the third pooling layer). It deals with the embedded space like a multivariate Gaussian, and it estimates the mean vector and covariance matrix for both real and generated data distributions, $\mu_r$, $\mu_g$, $\Sigma_r$ and $\Sigma_g$ respectively. The Fréchet Inception Distance is then calculated between the two Gaussians.
\begin{equation}
    \mathbf{FID}(r,g) = ||\mu_r - \mu_g ||_2^2 + Tr(\Sigma_r+\Sigma_g-2(\Sigma_r\Sigma_g)^{\frac{1}{2}})
\end{equation}
The trace function $Tr(.)$ is the sum of all the diagonal elements. Since it calculates how alike the authentic and generated distributions are, the lower the distance, the better the model; it is unlike the Inception Score, where the higher score is better. FID is usually a better metric than IS Since it is sensitive to GAN failures such as mode collapse. We explain failure modes in GAN's in section \ref{sec:stability}.

\section{Frequency Domain}
\label{sec:Freq}
Decomposing an image into different size scales with a basis function is a prominent idea in mathematics and engineering. The idea is to write any spatial signal as a sum of orthogonal basis functions to get a new representation based on the different resolutions of the original signal. This representation is known as frequency representation. Most of the decomposition techniques, including the ones described in this thesis, are bijective mappings, i.e., there is a unique representation in the frequency domain for each spatial signal. Arguably frequency domain's theoretical significance comes from the Convolution Theorem for standard Linear Time Invariant systems. The interactions of such systems with their input and output signals formulate by the convolution operator. The Fourier basis diagonalizes convolution, hence replace it with a simple element-wise multiplication in the frequency domain. The importance does not only summarize in theoretical works. The frequency-domain has opened up various tools for engineers. For audio signals applying filters to remove noise, detect different voices, or equalize the voices of different musical instruments are a small portion of many applications of frequency representations. In the case of the image spectrum, although the information is not apparent by the look of the spectrum, it contains valuable data related to the rate of change in moving to neighbor pixels, the lines and curves and edges of the objects, and noise in the images. It can also tell us how much information there is in a particular resolution, which is important when an image contains chaotic structures. There are helpful filters in noise reduction, edge detection, and many others operating on the images' frequency domain.\\
A few series and transforms have been introduced to change the representation of spatial periodic and non-periodic signals to a frequency space, such as Fourier transforms and series, Laplace transforms, and Z-transform. In this work, we adopted the arguably most famous decomposition technique, Discrete Fourier Transform (DFT).
\subsection{Power Spectrum}
A power spectrum is, in essence, a measure of the strength of the different features at different resolutions. Basically, We decompose an image into different size scales, for example, a coarser version like $2 \times 2$, and each new superpixel contains the dominant value of the corresponding pixels within the original image. If we subtract this dominant color from the original image, we are left with a version of the image without these largest structures. Repeating this procedure for less and less coarse versions of the original image give us a set of images (measurements on images) between the very coarse and the original image. Every image in between contains a measurement of how influential features in the original image are at that specific resolution, and the size scale associated with that resolution tells us what the size scale of those features is. For each resolution, the power spectrum is defined as the variance of the features on that resolution, i.e., how much total spread there is in values. A large spread means a strong signal and hence much variation on this scale.\\
To perform the size scale decomposition, we use DFT, which we will explain its detail in the next section. Basically, the time signal of each resolution corresponds to a sine and cosine pair with a period of $1/k$. We know $k$ as the frequency. Since we are performing a 2D transform, vector $\mathbf{k}$ consists of two elements as well $\mathbf{k}=(k_x, k_y)$ and each of the elements has two amplitudes for cosine and sine. As we discussed earlier, the power spectrum is the variance of the features in size scales. However, since different combination of $k$ elements can contribute to the same resolution (e.g. $\mathbf{k_1} = (k_x=1, k_y=2), \mathbf{k_2} = (k_x=2, k_y=1) $), the standard procedure is to bin them all together. The appropriate measure for the power for a specific $\mathbf{k}$ value is the average power within the corresponding $\mathbf{k}$ bin, multiplied with the volume of that bin in $\mathbf{k}$ space. Equation (\ref{eq:powerSpectrum}) shows the mathematical definition, and Fig \ref{eq:powerSpectrum} shows an example of the spectrum, power spectrum pair, indicated by two different $\mathbf{k}$ bins and their corresponding location on both spectrum and power spectrum.

\begin{equation}
    PS(r) = \frac{1}{2\pi} \int_0^{2\pi} |F(r, \theta)| d\theta
    \label{eq:powerSpectrum}
\end{equation}
where $r$ and $\theta$ are coming from the standard change from Cartesian to polar coordinate change:
\begin{equation*}
    F(k,l) = F(r, \theta): \quad r = \sqrt{k^2+l^2}, \quad \theta = \tan^{-1}(l,k)
\end{equation*}

\begin{figure}[ht]
\caption{ left: spectrum of an image, right: power spectrum of the same image. yellow and red circles are equivalent resolutions which binned together in the power spectrum. Figure credit for \cite{Durall_2020}}
\centering
\includegraphics[width=0.8\textwidth]{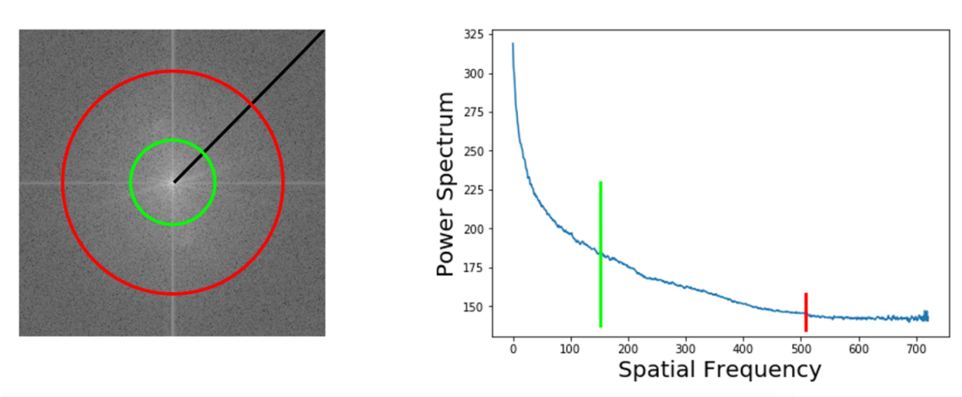}
\label{fig:powerSpectrum}
\end{figure}

\subsection{Discrete Fourier Transform (DFT) and Fast Fourier Transform (FFT)} 
Fourier Transform presumably is the most common signal decomposition in the literature, and it is the foundation of the harmonic analysis. DFT is a sampled version of the Fourier transform, so it does not contain all the frequencies in the spatial function. however, it contains enough frequency elements to describe one and only one spatial function without losing any information. 1D-DFT of the the function $f(x)$ with $N$ points $N = 0, 1, ..., N-1$ is calculated as follow:
\begin{equation}
    F(k)=\frac{1}{\sqrt{N}}\sum_{i=0}^{N-1} f(x) e^{-i2\pi\frac{kx}{N}}
\end{equation}
To get back to the spatial domain the inverse DFT is as follow:
\begin{equation}
    f(a)=\frac{1}{\sqrt{N}}\sum_{k=0}^{N-1} F(k) e^{i2\pi\frac{ka}{N}}
\end{equation}
The exponential terms are the basis function corresponding to each point in the Fourier domain representation $F(k)$. The value of each point in $F(k)$ is calculated by multiplying each point in the spatial domain with the corresponding Fourier basis and then summing them all up. In a more formal fashion, DFT expresses the spatial function $f(x)$ as a linear combination of orthogonal oscillating basis functions $e^{-i2\pi\frac{kx}{N}}$, indexed by their rate of oscillation (frequency).\\
Dealing with the images, we need to extend the 1D-DFT to 2D-DFT. Given an $N\times N$ square image $f(i,j)$ , we can extend the definition of DFT and iDFT to two dimensions as follow:
\begin{equation}
    F(k,l)=\frac{1}{N}\sum_{i=0}^{N-1} \sum_{j=0}^{N-1} f(i,j) e^{-i2\pi(\frac{ki}{N}+\frac{lj}{N})}
\end{equation}
\begin{equation}
    f(a,b)=\frac{1}{N}\sum_{k=0}^{N-1} \sum_{l=0}^{N-1} F(k,l) e^{i2\pi(\frac{ka}{N}+\frac{lb}{N})}
\end{equation}
Note that the regularization constant $\frac{1}{N^2}$ only appears in the inverse function. The N is usually a power of 2, and when the original function does not satisfy the size, it is usually zero-padded to get to the desired size. DFT output is a complex function, and they can be shown with two different images, as \textit{real} and \textit{imaginary} or \textit{magnitude} and \textit{phase}. Neither part can be ignored as both of them contain information about the original image.\\
The complexity of the DFT computation is $O(N^2)$; however, with a computational trick, it can be reduced to $O(N\log_2 N)$, the algorithms calculating the DFT with reduced complexity are known as Fast Fourier Transform (FFT). We use the terms DFT and FFT interchangeably in this document as the output to them is the same, only the procedure to get the output is different.

\begin{figure}[ht]
\caption{ Demonstration of decomposing a signal into different cosine elements }
\centering
\includegraphics[width=0.8\textwidth]{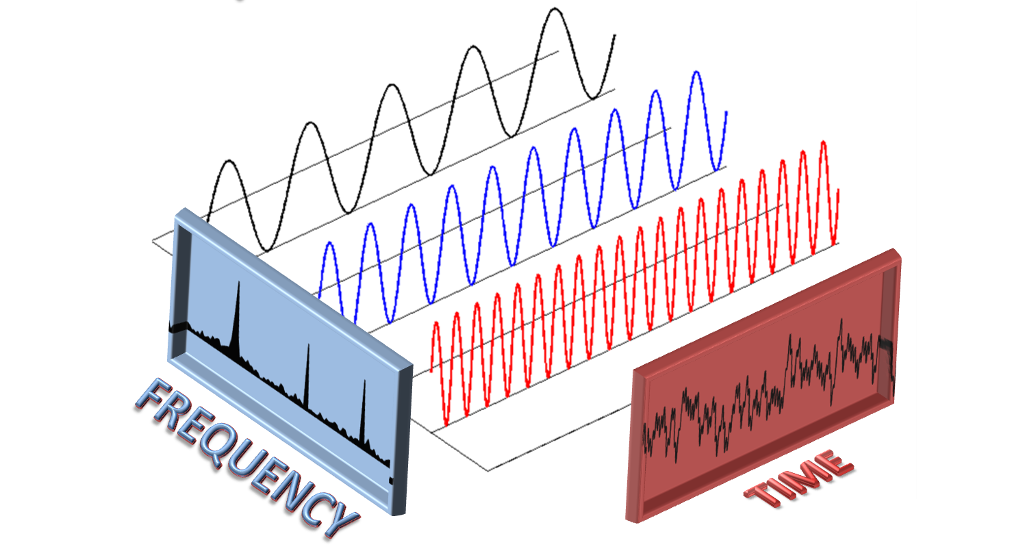}
\label{fig:freqAndTime}
\end{figure}
    \chapter{Literature Review}
\label{chap:literatureReview}
Since \cite{GAN_original} introduced GANs in 2014, there has been a considerable body of work on improving and stabilizing this category of neural networks. \cite{DCGAN} proposed a set of constraints, making GAN training with convolutional layers more stable. However, they could not overcome the mode collapse problem in training. They also showed the vector arithmetic properties of the latent input space of the generator similar to the known relations in the word embedding. Their other contribution was showing that adversarial network pairs in GANs can learn meaningful representation from objects to scenes. \cite{wgan, wgan_gp} brought up a new distance to measure the difference between original data and generated data distributions, leading to the most noteworthy GAN success due to robustness to architectural choices. They were the first to address the mode collapse problem significantly. It is also worthy of mentioning that their new distance (estimated EM distance) provided meaningful learning curves for debugging and hyperparameter tuning.  \cite{sngan} applied spectral normalization over the weight matrix, which not only helped the stability of the discriminator network but also, due to its significant computational benefits over the previous methods, made generating high-quality class conditional ImageNet \cite{imagenet} scale training possible. \cite{sagan} used self-attention in both generator and discriminator, leading to a significant increase to the inception score of the GANs as well as delivering high-quality unconditional full-ImageNet samples. Their method involved a self-attention procedure allowing both generator and discriminator to have access to non-local regions of images when trying to detect or generate images. \cite{bigGAN} continued the efforts to scale up the dataset with a new variant of orthogonal regularization and truncation trick. Their model allows increasing the batch size and model width, which not only allows for a significant scale-up but also increases the performance in the Inception Score. \cite{bigan, bigbigan} modified the discriminator by adding an extra path for the encoded generated images to decide not only based on the data space but also by another representation of the latent space. They introduced a novel representation learning approach to train GANs, which was one of our motivations in the current work.
StyleGAN based GANs \cite{styleGAN, styleGAN2} at the time of writing this document is holding the state-of-the-art in the literature. Their main contribution was the new architecture for the generator where it does not feed noise $z$ directly to the generator. instead, it passes through an MLP to get a style vector $w$ and feed in the generator architecture at different layers.\\
Although GANs have succeeded in delivering high-quality images almost impossible for human eyes to detect, recent studies \cite{Zhang_2019, easytospot, leveraging, Durall_2020} showed that it is not an arduous task to detect GAN-generated images. In the core of the forensic analysis of \cite{Zhang_2019, leveraging,Durall_2020}, lies the discrepancies between the spectrum of the generated images vs. the real ones. To further relate the frequency notion to the current state of neural network training in the literature \cite{spectralbias} found evidence of a spectral bias on neural networks, following this idea \cite{bandlimited} came up with band-limited training of the CNNs. Another frequency evidence lies in \cite{jiang} which provided theoretical proof on having band-limited model as a condition for perfect learning in neural networks.\\
 Learning in the frequency domain is not well-investigated in the literature. However, as we explained above, the clues are further pushing us to take the frequency domain More seriously. currently, the models are using fully connected \cite{frequency0} or CNN-based architectures\cite{frequency1} in the input. However, the recent CVPR paper \cite{frequency2} proposed a model based on DCT transformation, which aims to retain more original picture information through DCT transformation and reduce the communication bandwidth between CPU and GPU. In this work, we use geometric deep learning \cite{geometric} to introduce a new model designed to perform specifically on the frequency domain of the image. We will explain the related literature review on geometric deep learning in more detail in the chapter \ref{chap:GeometricDeepLearning}.

    \chapter{Geometric Deep Learning}
\label{chap:GeometricDeepLearning}
In this chapter, we give a brief overview of geometric deep learning \cite{geometric} and how it brings together a large class of neural networks working on different data types such as unstructured sets, grids, graphs, and manifolds to interpret under unified principles. Then we describe symmetries in the image data form and how geometric deep learning uses them to add inductive bias to the neural network framework, reducing the search space in the hypothesis class for more suitable functions. In section \ref{freqModule} we use the idea of equivariant functions to devise a suitable deep neural network architecture for the frequency domain. Our proposed architecture searches the frequency while keeps the essential property of shift invariancy of the image in mind. To the best of the author's knowledge, a systematic architectural design for the frequency domain has never been done in the literature. We consider it one of our main contributions.

\section{Inductive bias}
\label{sec:priors}
In machine learning theory, supervised learning is usually formalized by assuming that labels $y$ are generated by an unknown function $f$, such that $y_i = f(x_i)$.  Also, let us assume that the data points and labels come from an i.i.d sampling of the underlying data. $\mathcal{D}=\{(x_i, y_i)\}_{i=1}^N$, where $N$ here is the number of observations. The learning objective here is to estimate the function $f$ as best as possible from a set of parameterized functions we have, known as a hypothesis class, $\mathcal{H}=\{f_{\theta\in\Theta}\}$. Neural network architecture can be seen as an instance of such parameterized hypothesis classes with parameters $\theta\in\Theta$ as the networks' weights. The ideal method to evaluate the chosen function from the hypothesis class is the expected loss defined as follow:
\begin{equation*}
    \mathcal{L}_P(f^*):= \mathbb{E}_P\{L(\Tilde{f}(x),f(x)\}
\end{equation*}
$\Tilde{f}\in\mathcal{H}$ is the chosen function, and $P$ is the underlying distribution of the data. $L(.,.)$ is the suitable loss for our task; some examples can be square loss and binary cross-entropy. Thus, to have successful learning, one needs to consider a notion of regularity in choosing functions $\Tilde{f}$. This idea of regularization is known as the inductive bias, which can be applied via various methods, explained further in this section.\\
To have a successful learning scheme, the other important and even more prominent property is that the hypothesis class should be dense enough to have the capacity to approximate functions over current large-scale, high-quality datasets. However, in the case of neural networks, it has never been a problem. Even a simple two-layer perceptron, $f(\bx) = \mathbf{c}^\intercal \sigma(\mathbf{A}\bx+\mathbf{b})$ shown to be dense in the space of continuous functions on $\mathbb{R}^d$.\\
To formalize the regularity notion into the learning scheme, we can get help from complexity measure $c : \mathcal{H} \to \mathbb{R}_+ $, now the problem can be seen as:
\begin{equation*}
    \Tilde{f}\in \arg \min_{g\in\mathcal{H}} c(g) \quad \text{s.t.} \quad g(x_i) = f(x_i) \quad \text{for} \quad i=1,...,N
\end{equation*}
The realizability assumption is considered for this formulation for convenience. In other words, now we are looking for the least complex (more regular) function among our hypothesis class. For the complexity measure, different norms of the working space are a well-known option used in learning approaches such as logistic regression or SVMs. In neural networks, network weights are a proper choice for complexity measurements. i.e. $c(f_\theta)=c(\theta)$. While explicitly applying regularities to the empirical loss is a common, well-defined approach known as "Structural Risk Minimization," the Implicit approaches are getting more and more attention of researchers \cite{implicit1, implicit2}.\\
As the dimension of the space goes higher, finding suitable regular functions will get more challenging as well. In the case of the long-established regularity notion, 1-Lipschitz functions. The number of observations needed for the desired loss will grow exponentially. The good news is that real-world applications such as computer vision tend to work with functions that inherently possess some spatial structures in their physical domains. The geometric deep learning objective is to exploit structures as a notion of regularity further to reduce the search space for the most proper function.

\section{Symmetries as Regularities}
\label{sec:symmetry}
Symmetries on a domain or a signal are transformations that leave certain properties of the domain of the signal, or the signal itself, unchanged. Real-world applications are full of these operations. In computer vision's object detection tasks, shift or rotation are examples of transformations that leave the type of object invariant. If there is an image of a rhinoceros, the shifted or rotated rhinoceros should still be classified as a rhinoceros.\\
The symmetries of the domain $\Omega$, and signal $\mathcal{X}(\Omega)$ operating on it present a constraint on the function $f$ operating on those signals. It turns out that these symmetries are powerful prior to consider as inductive bias. They will reduce the search space $\mathcal{H}(\mathcal{X}(\Omega))$ to only the functions satisfying the invariancy. Here we explain the two groups of functions we use their symmetric properties in this thesis, \textit{invariants} and \textit{equivariants}.

\paragraph{Invariant functions:}
A function $f: \mathcal{X}(\Omega) \to \mathcal{Y}$ is invariant with respect to transformation $\textgoth{g}$ if $f(\rho(\textgoth{g})x)=f(x)$ and $x\in \mathcal{X}(\Omega)$, in other words, if its output is unchanged after the transformation.\\
Assuming the transformation $\textgoth{g}$ is linear i.e $\textgoth{g}.(\alpha x+\beta x^\prime) = \alpha(\textgoth{g}.x)+\beta(\textgoth{g}.x^\prime)$, $\rho(g)$ is the matrix equivalent of the the transformation $\textgoth{g}$.
The function for the rhinoceros classification example is categorized under shift invariant functions, which is very common in pattern recognition literature. 
Classic perceptron networks hypothesis' class consists of functions not having the shift invariant property. It is making them an unseemly choice for vision tasks where the object's position is not relevant in the final result. However, as we explain later in section \ref{sec:CNNwGDL}, applying a geometric prior for the shift would result in an architecture known as Convolutional Neural Networks (CNNs) (for further reading on CNNs please refer to section \ref{sec:NN}). That being said, shift geometric prior in the CNNs is not shift invariancy. In fact, it is shift equivariancy. That is to say, a particular amount of shift in the input would result in the same amount of shift in the output. Formally we can define an equivariant function as follow:
\paragraph{Equivariant functions:} A function $f: \mathcal{X}(\Omega) \to \mathcal{Y}$ is equivariant with respect to transformation $\textgoth{g}$ if $f(\rho(\textgoth{g})x)=\rho(\textgoth{g})f(x)$ and $x\in \mathcal{X}(\Omega)$. In other words, If the output react in the exact same way input reacted to the transformation.\\
A computer vision example of such tasks is image segmentation. If the segmented input moved, we expect the output to shift the same amount. 
Another noteworthy property of the invariant and equivariant function arises in their combinations. If $f_1(x)$ and $f_2(x)$ are equivariant and $g_1(x)$ and $g_2(x)$ are invariant, then by definition, $f_1 \circ f_2 (x)$, $g_1 \circ g_2 (x)$ and $f_1 \circ g_1 (x)$ are equivariant, invariant and equivariant respectively. We also know the following properties for symmetric transformation, by the axioms of symmetric transformations.\\
\textbf{Associativity:} For all the functions belong to the same symmetry group $f_1 \circ f_2 (.) = f_2 \circ f_1 (.)$. Although invariants and equivariant are not from the same symmetry group, by definition they are also have this property together.\\
\textbf{Inverse:} for every function belong to the same symmetry group there is a unique inverse in the same symmetry group, e.g. $g_1^{-1} \circ g_1(x)=x$\\
Geometric deep learning provides a blueprint for constructing specific architectures satisfying the explained conditions. For the specific case of equivariant functions. the blueprint's building blocks are as follow:\\
\textbf{Linear equivariant layer $B$:} $\mathcal{X}(\Omega) \to \mathcal{X}(\Omega)$\\
\textbf{Non-linearity $\sigma$:} applied element-wise  as $(\sigma(x))(u) = \sigma(x(u))$.\\
\textbf{Local pooling (coarsening) $P$:} $\mathcal{X}(\Omega) \to \mathcal{X}(\Omega)$\\
\textbf{Invariant global pooling layer $A$:} $\mathcal{X}(\Omega) \to \mathcal{Y}$\\
By using the mentioned building blocks in cascade, we can construct the equivariant function f as follow:
\begin{equation*}
    f = A \circ \sigma_J \circ B_J \circ P_{J-1} \circ ... \circ P_1 \circ \sigma_1 \circ B_1
\end{equation*}
The building blocks are designed such that the output of each block matches the input space of the next layer.
\begin{figure}[ht]
\caption{The blueprint for designing architectures using geometric priors (image credit for: \cite{geometric})}
\centering
\includegraphics[width=0.8\textwidth]{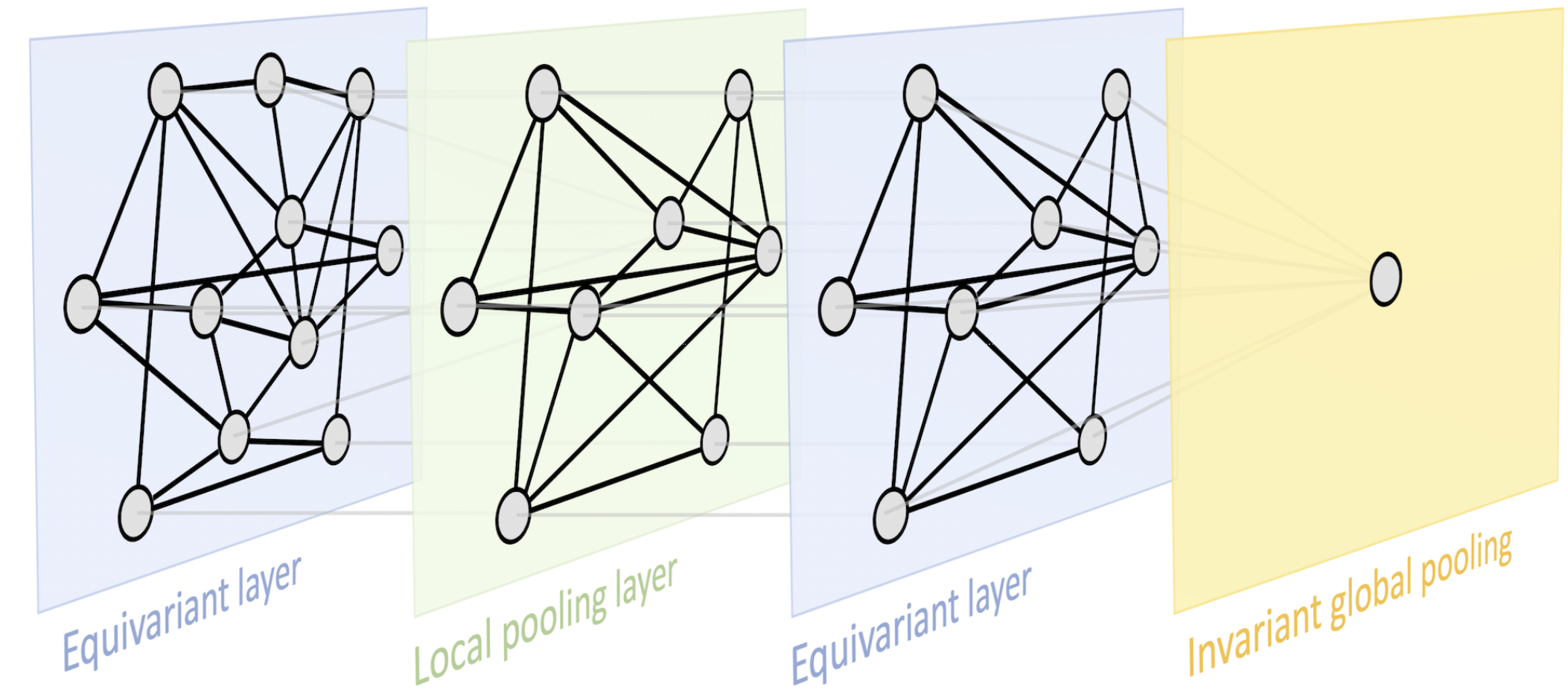}
\label{fig:blueprint}
\end{figure}
\section{Designing CNNs with GDL:}
\label{sec:CNNwGDL}
This section describes how we can apply the blueprint to a homogeneous grid space, where images are defined as signals over this space. Understanding this design is essential for the cause of this thesis since designing the Fourier layer for the discriminator of the GAN uses the same method with a similar approach.\\
Let's assume that we have a one dimensional grid for simplicity. in this grid each element $\bx_u$ has a right and a left neighbor $\bx_{u-1}, \bx_{u+1}$. Using the blueprint we can design the equivariant function $\mathbf{F}$ with a local aggregation function $\phi$ operating over a grid element and its neighbors $\phi(\bx_{u-1}, \bx_u, \bx_{u+1})$. Choosing $\phi(\bx_{u-1}, \bx_u, \bx_{u+1}) = \theta_{-1} \bx_{u-1} + \theta_0 \bx_u + \theta_1 \bx_{u+1}$ yields the function $\mathbf{F}$ as the matrix product:
\begin{equation*}
    \mathbf{F}(\mathbf{X}) =
    \begin{bmatrix}
    \theta_0 & \theta_1 & & & \theta_{-1}\\
    \theta_{-1} & \theta_0 & \theta_1 & &\\
    & \ddots & \ddots & \ddots & \\
    & & \theta_{-1} & \theta_0 & \theta_1\\
    \theta_1 & & & \theta_{-1} & \theta_0
    \end{bmatrix}
    \begin{bmatrix}
    -- & \bx_0 & --\\
    -- & \bx_1 & -- \\
     & \vdots &  \\
    -- & \bx_{n-1} & -- \\
    -- & \bx_n & -- 
    \end{bmatrix}
\end{equation*}

In the machine learning, using this form of diagonal matrix is known as weight sharing. This matrix in an example of a circulant matrix, which is essentially a vector $\mathbf{\boldsymbol{\theta}} = (\theta_0, \dots, \theta_{n-1})$ append to itself after circular shifts. $\mathbf{C(\boldsymbol{\theta})}=(\theta_{u-v\: mod \:n})$. product of a vector with a circulant matrix is equivalent with cyclic discrete convolution.
\begin{equation*}
    \mathbf{C(\boldsymbol{\theta})}\bx = (\bx*\boldsymbol{\theta})_u = \sum_{v=0}^{n-1} x_{v \: mod \: n} \: \theta_{u-v\: mod \:n}
\end{equation*}
In machine learning and signal processing, the above $\boldsymbol{\theta}$ is known as the filter, and its weights/values are subject to the learning process. Circulant matrices have commutativity property, which means their product is commutative, and it is also circulant. More mathematically $\mathbf{C(\boldsymbol{\theta})C(\boldsymbol{\eta})}=\mathbf{C(\boldsymbol{\eta})C(\boldsymbol{\theta})}$ 
If we choose $\boldsymbol{\theta} = (0,1,0,\dots,0)^\intercal$ the resulted circulant matrix $\mathbf{S}$, shift vectors one position right (circularly) and we call it the shift matrix. From the properties of the convolution, and circulant matrices, we get the desired shift equivariance:
\begin{equation*}
    \mathbf{SC(\boldsymbol{\theta})x} = \mathbf{C(\boldsymbol{\theta})Sx}
\end{equation*}
An interesting fact is that the other way around is also true. A matrix is circulant if it commutes with the shift. In other words, convolutions are the only linear operators keeping the equivariancy.
    \chapter{Methodology}
\label{chap:ProblemDef_Methodology}
\section{Frequency in the Neural Networks}
\label{sec:FreqGAN}
\begin{figure}[ht]
\caption{transposed convolution and nearest neighbour up-sampling (image credit for \cite{Zhang_2019})}
\centering
\includegraphics[width=0.8\textwidth]{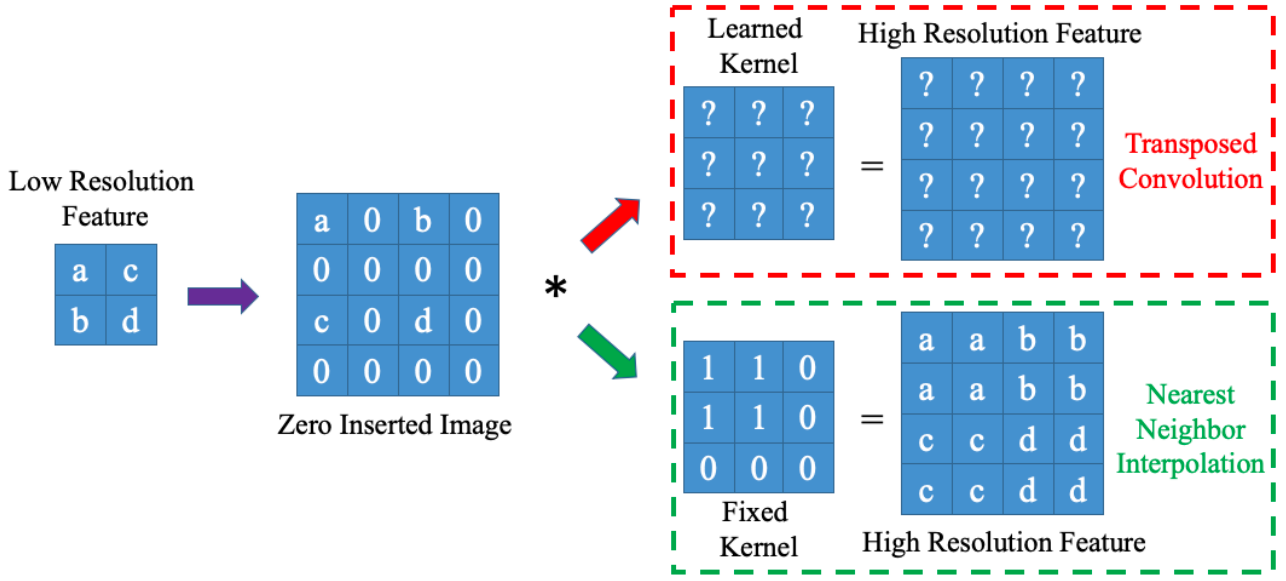}
\label{fig:conv}
\end{figure}

\begin{figure}[ht]
\caption{Left: spectrum of the real image; Right: spectrum of the generated image, the checkerboard artifacts are zoomed in, the extra high frequency components are visible in spectrum of the fake image. (image credit for \cite{Zhang_2019})}
\centering
\includegraphics[width=0.8\textwidth]{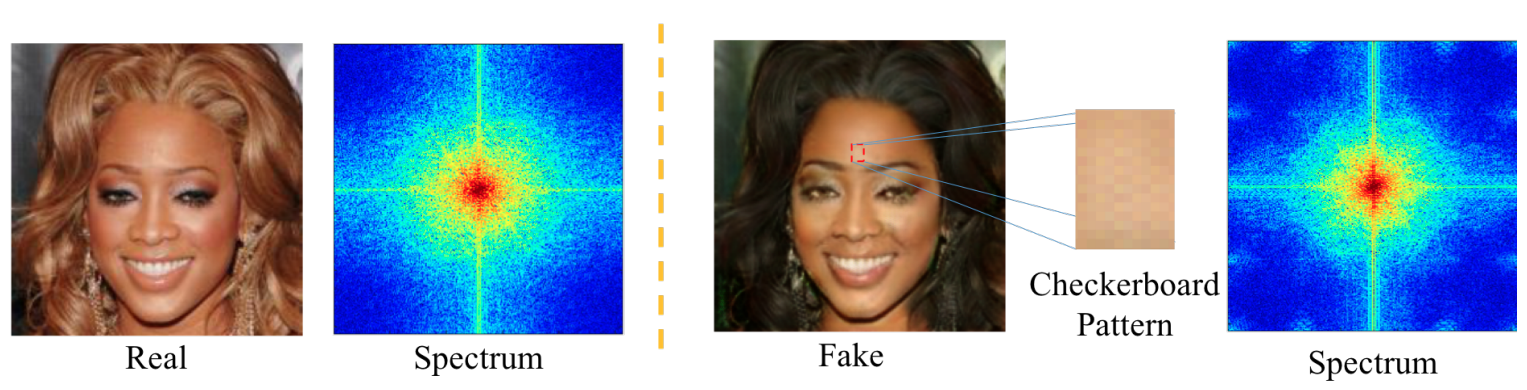}
\label{fig:artifacts}
\end{figure}

Recent state-of-the-art GAN architectures used for vision tasks utilized the convolutional neural networks (CNN) structure for their generator and discriminator modules \cite{styleGAN, styleGAN2, bigGAN}. Regardless of the network architecture and the loss function difference, they all need to map from a low dimensional latent space to a high dimensional high-resolution image space. To this end different GANs used different upsampling approaches like bi-linear \cite{styleGAN, styleGAN2, bigGAN}, transposed convolution \cite{DCGAN, DRAGAN, LSGAN} or nearest neighbor \cite{karras2017progressive, Park_2019}. In the simple case, when the up-sampler scales the image by a factor of $m=2$; it inserts a zero between all the pixels in each row/column. Then it applies the convolution to interpolates the inserted zeros with suitable values. A simple formulation of the last two is shown in Figure \ref{fig:conv}.
\cite{Zhang_2019} first analyzed the frequency consequences of the GAN's up-sampling modules. Properties of the DFT (Discrete Fourier Transform) reveal that zero-insertion in an image will affect the spectrum in a way that looks like multiple copies of the original spectrum attached together. The proof for the one-dimensional case is as follows, and it is easily extendable to the two-dimensional case. Given $x(n), n=0,...,N_1$ as the input signal to the up-sampler and $X(k), k=0,...,N-1$ as its DFT. By inserting the zeros we'll get $x'(n), n=0,...,2N-1$ where $x(2n)=x(n)$ and $x(2n+1)=0$ for $n=0,...,N-1$. Now the DFT of the $x'(n)$ will be $X'(k)$. For $k<N$,
\begin{multline}
    X'(k) = \sum_{n=0}^{2N-1} x'(n) exp(\frac{-i2\pi}{2N}kn) =\\ \sum_{n=0}^{N-1} x(n) exp(\frac{-i2\pi}{2N}k(2n)) = X(k)
\end{multline}
for $k \geq N$, let $k'=k-N$, thus $k' = 0,...,N-1$ then,
\begin{multline}
        X'(k) = \sum_{n=0}^{N-1} x(n) exp(\frac{-i2\pi}{2N}(k'+N)(2n))=\\ \sum_{n=0}^{N-1} (x(n) exp(\frac{-i2\pi}{N}nk'-i2n\pi))=X(k')
\end{multline}
The above shows that there will be two copies of the spectrum of the $X(k)$ in $X'(k)$ first at $[0,...,N-1]$ and second at $[N,...,2N-1]$. This transformation introduces high-frequency components to the spectrum. \cite{odena2016deconvolution} discussed that these high-frequency components play a significant role in a checkerboard artifact phenomenon seen in the outputs of generative convolutional networks. Figure \ref{fig:artifacts} demonstrate the high-frequency components in the spectrum as well as the checkerboard artifacts. In the next step of the up-sampling, when the convolution occurs, we have neither a guarantee for the learned convolution kernel to be a suitable low-pass filter nor a sampling strategy to avoid such high-frequency artifact appear in the subsequent layer output. Thus, it can not fade out the newly introduced high-frequency components.

\section{DFT and Principles of Symmetry}
\label{sec:DFTGDL}
Maybe the most famous property of the Fourier domain in signal processing arises from the convolution theorem. That is to say, the convolution operation in the spatial domain is equivalent to element-wise multiplication in the Fourier domain. It is important since the interactions of linear time invariant (LTI) signals and systems can be described with the convolution operation. This property is derivable with properties of symmetry, and circulant matrices explained in section \ref{sec:CNNwGDL}. To that end, we need to remember that diagonalizable matrices share common eigenvectors (with different eigenvalues) iff they mutually commute. Since circulant matrices satisfy these conditions, we can calculate the eigenvectors of one of them. Shift matrix $\mathbf{S}$ is a convenient choice here, whose eigenvectors luckily happens to be the DFT basis:
\begin{equation*}
    \boldsymbol{\phi}_k = \frac{1}{\sqrt{N}} \left( 1, e^{\frac{2\pi ik}{N}}, e^{\frac{4\pi ik}{N}}, \dots, e^{\frac{2\pi i (n-1) k}{N}}  \right)^\intercal \quad , \qquad k=0,1,\dots, n-1 
\end{equation*}
Arranging the $\boldsymbol{\phi}$ vectors into a $n \times n$ matrix gives us $\boldsymbol{\Phi} = (\boldsymbol{\phi}_0, \dots, \boldsymbol{\phi}_{n-1})$. Matrix multiplication with $\boldsymbol \Phi$ yield the inverse DFT and with its complex conjugate $\boldsymbol{\Phi^*}$ the DFT.
All circulant matrices share this eigenvectors, so the Fourier transform of the filter is the eigenvector matrix $\fo(\boldsymbol{\theta}) = \boldsymbol{\Phi}^*\boldsymbol{\theta}$ for the circulant matrix $\boldsymbol{C}(\boldsymbol{\theta})$. Thus, we can write $\boldsymbol{C}(\boldsymbol{\theta})$ in diagonalized form to obtain the convolution theorem:
\begin{equation*}
    \boldsymbol{C}(\boldsymbol{\theta})\bx =\boldsymbol{\Phi} 
    \begin{bmatrix}
    \fo(\theta_0) & &\\
     & \ddots & \\
     & & \fo(\theta_{n-1})
    \end{bmatrix}
    \boldsymbol{\Phi}^* \bx = \boldsymbol{\Phi} \left( \fo(\boldsymbol{\theta}) \odot \fo(\bx) \right)
\end{equation*}
where $\odot$ is element-wise multiplication. we use the above mentioned characteristic of circulant matrices to provide a proof of work for our frequency domain architecture in the next section.

\section{Problem Definition}
\label{sec:ProblemDef}
\cite{Zhang_2019} shed light on the dissimilarities between authentic image spectrum and the images generated by convolutional networks. These disparities mostly show themselves in the forms of high-frequency patches in the generated images spectrum, discussed by \cite{odena2016deconvolution} as responsible for the checkerboard artifacts showed in them. Knowing that Fourier is a bijective mapping, any apparent difference in the spectrum demonstrates a flaw in learning the original distribution, which is the main objective of training GANs as a generative model. In this work, we are developing a new architecture for the discriminator to address the spectral difference where the discriminator gets the images in the Fourier and the spatial domains. We then experimentally demonstrate that our model can enhance the learned distribution, emphasizing the importance of the spectrum of images on the learning process.
\section{Methodology}
\label{sec:methodology}
The most straightforward way for solving dissimilarities in the spectrum is to add frequency information to the discriminator. \cite{pix2pix, bigan, bigbigan} have already proved that feeding information to the discriminator can boost the performance of the GANs for other applications. In this case, the network would better detect high-frequency components' differences in the generated images by providing the discriminator with frequency information. Thus, it pushes the generator to produce images matching the actual image distribution's spatial and frequency representation. Equation (\ref{eq:FreqGAN1}) shows the new loss function for the GAN incorporating the Fourier transform $\mathcal{F}(.)$ of the images directly in the loss.
\begin{multline}
        \min_\cG \max_\cD V(\cD,\cG) =\\ \mathbb{E}_{\bx\sim P_{\bx}(\bx)}[\log \cD(\bx, \fo(\bx))]+ \mathbb{E}_{\bz\sim P_{\bz}(\bz)} [\log (1-\cD(\cG(\bz),\fo(\cG(\bz))))].
        \label{eq:FreqGAN1}
\end{multline}
For preprocessing the Fourier information \cite{Durall_2020} discusses the following 1D representation of Fourier transform via azimuthal integration over radial frequencies, practical enough to point out the dissimilarities of the spectrum and spatial domain. Assuming that images are $M \times M$.
\begin{equation}
    AI(\omega_k) = \int_0^{2\pi} \norm{\fo(\omega_k \cdot \cos{(\phi)}, \omega_k \cdot \sin{(\phi)})}^2 d\phi
    \label{AItegral}
\end{equation}
\begin{equation*}
    \mathrm{for} \qquad k=0, ..., \frac{M}{2}-1
\end{equation*}
Equation (\ref{AItegral}) can be seen as a coordinate conversion from Cartesian to polar and then getting the mean intensity over the radial distance. It is also used in \cite{ssdgan} as the input to the frequency module. Despite its popularity due to more straightforward computation in 1D, performing average operation will cause loss of information in FFT amplitudes and completely disregard the phase information, which proved to be an essential part of the spectrum \cite{phase_importance1, phase_importance2}. In this work, we use both amplitude and phase of the image as the input to our frequency module to fully exploit the spectrum information.\\

%% revisions 
At a high level, our work consists of two sections. First, we introduce a high-level architecture with unary losses (Figure \ref{fig:architecture}) to incorporate frequency information into the GAN architecture in a guided manner. It will result in a discriminator more sensitive to frequency discrepancies, which pushes the generator to produce images more realistic in the frequency domain. Second, we devise a Frequency module to use in our architecture. This module's task is to get the frequency spectrum and output a realness score. The frequency module uses a base layer we named \textit{EV-Freq} which is designed based on the blueprint provided by geometric deep learning.
%%%%%%%%%%%
\subsection{The High-Level Architecture}
\begin{figure}[ht]
\caption{Architecture for FreqGAN}
\centering
\includegraphics[width=0.8\textwidth]{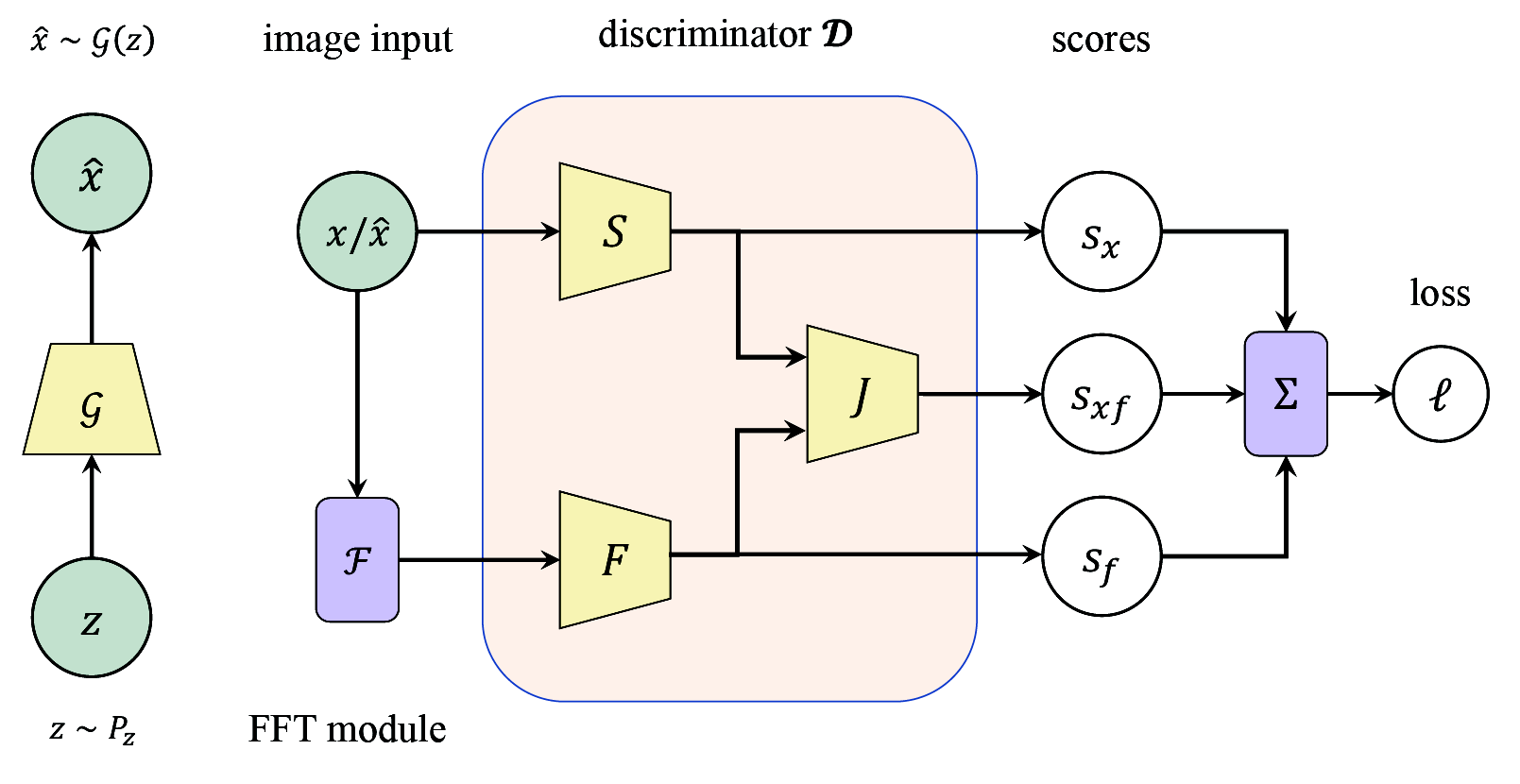}
\label{fig:architecture}
\end{figure}

Importing the spectrum to the network has shown degradation in the performance of the model \cite{ssdgan}. This phenomenon is a consequence of losing too much image information in the spectrum, hence, not having enough gradient for the adversarial training. In order to overcome this problem, we adopt the three unary terms proposed by \cite{bigbigan} to guide the optimization.\\
Figure \ref{fig:architecture} shows a high-level view of our model's architecture. The generator and discriminator loss in our model can be described as follow:
\begin{align*}
    s_\bx(\bx) &= \theta_\bx^\intercal S_\Theta(\bx) \\
    s_f(\bx) &= \theta_{f}^\intercal F_\Theta(\fo(\bx))\\
    s_{\bx f}(\bx) &= \theta_{\bx f}^\intercal J_\Theta(S_\Theta(\bx),F_\Theta(\fo(\bx)))\\
    \ell_\cG(\bx,y) &= y(s_\bx(\bx)+s_f(\bx)+s_{\bx f}(\bx)) \qquad y\in \{ -1, +1 \} \\
    \cL_\cG(P_\bx, P_\bz) &= \mathbb{E}_{\bx\sim P_\bx}[\ell_\cG(\bx,+1)]+\mathbb{E}_{\bz\sim P_\bz, \hat{\bx}\sim \cG_\Phi(\bz)}[\ell_\cG(\hat{\bx},-1)]\\
    \ell_\cD(\bx,y) &= \ell(ys_\bx(\bx))+\ell(ys_f(\bx))+\ell(ys_{\bx f}(\bx)) \qquad y\in \{ -1, +1 \} \\
    \cL_\cD(P_\bx, P_\bz) &= \mathbb{E}_{\bx\sim P_\bx}[\ell_\cD(\bx,+1)]+\mathbb{E}_{\bz\sim P_\bz, \hat{\bx}\sim \cG_\Phi(\bz)}[\ell_\cD(\hat{\bx},-1)]
\end{align*}
$\ell$ in the sixth line refers to the loss we chose to apply on the instances. \cite{bigbigan} chose hinge loss for this, and we intend to do the same here. The discriminator $\cD$ has three modules. $S$ takes the images in the data $\bx$ only in the spatial domain while $F$'s input is the data in the frequency domain $\fo(\bx)$. With this model, we get two separate scores for the realness of the image's spectrum and images in the spatial domain. Each module calculates a realness scalar score based on its input. The scalar scores respectively are $s_\bx$ and $s_f$ calculated with their parameters $\theta_\bx$ and $\theta_f$. The desirable output should get a high score both in frequency and spatial domains. Thus, the last module $J$ gets the output of the other two modules,  $s_\bx$ and $s_f$, as its inputs, and the output is the joint score $s_{\bx f}$ which defines how real the image is concerning both discussed domains. The generator's $\cG$ parameters $\Phi$ and discriminator's $\cD$ are optimized to minimize the losses $\cL_\cG$ and $\cL_\cD$ respectively. The Discriminator loss $\cL_\cD$ trains the discriminator to identify the joint representations of its input coming from real data, $(\bx, \fo(\bx))$, or the generator, $(\cG(\bz), \fo(\cG(\bz)))$ with predicting, respectively, positive and negative values for them.
%% revision
The discriminator Loss consists of three losses coming from spatial, frequency, and joint modules. Each of them is responsible for $1/3$ of the total loss. We recognize the equal weights for each module as a limitation for our work, and we planed to design an architecture with adaptive weights in our future works.
The generator loss, $\cL_\cG$ optimize the generator to misguide the discriminator to predict incorrectly, stirring the generator to create images matching both representations (in spatial and frequency domain) of the real data. \\

\subsection{Frequency Module}
\label{freqModule}
\begin{figure}[H]
\caption{a) our designed function operating over and image b) desired behaviour of our designed input over shifted input}
\centering
\includegraphics[width=0.8\textwidth]{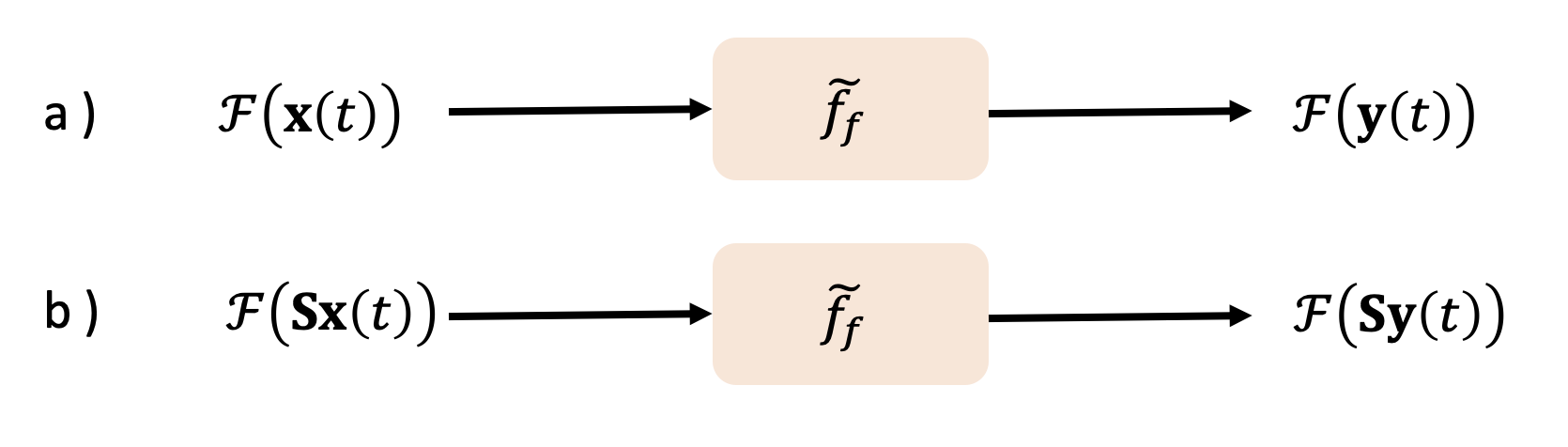}
\label{fig:shiftEquivariance}
\end{figure}

Deep learning researchers developed various deep neural network architectures for different spaces and signals defined over them, such as sets, graphs, images, and text. However, there is no particular design for the frequency spectrum. Current frequency-driven neural networks use fully connected or convolutional neural networks. The former results in a vast search space for optimization. Therefore, a slight chance to exploit the approximation function over frequency. The latter, without any modification, is designed for spatial feature extraction, which is by itself not a helpful feature in the spectrum data. Geometric deep learning establishes a framework for unifying some of the well-defined architectures. Moreover, it provides a blueprint for systematically design a new architecture for new domains and signals. In this section, by following the GDL blueprint, we propose a new architecture EV-Freq for frequency-domain signals.
In section \ref{sec:CNNwGDL}, shift equivariancy of the image's spatial domain provided the inductive bias needed for designing CNNs. We use the same idea here. Our functions in the frequency domain should not disturb the equivariancy of their corresponding spatial images. Let's denote out input image with $\bx(t)$, our chosen function operating over frequency domain from our designed hypothesis class as $\Tilde{f_f} \in \mathcal{H}_f$, and also $\fo \left( \mathbf{y}(t) \right) = \Tilde{f_f}\left( \fo \left( \bx (t) \right) \right)$. In other words, $\mathbf{y}(t)$ is the Fourier inverse of the $\Tilde{f_f}$ output. Figure \ref{fig:shiftEquivariance}-a shows the relationship. \\
As we discussed in section \ref{sec:DFTGDL}, we can make the Fourier transfer of a function by matrix multiplication with matrix $\boldsymbol{\Phi}^*$. We use this notation in the following proof whenever we perform a Fourier transform on a signal.
Aimed at keeping the equivariance in the spatial domain $\Tilde{f_f}$ should satisfy the following equation: (Figure \ref{fig:shiftEquivariance}-b) 

\begin{equation}
    \Tilde{f_f} \left( \boldsymbol{\Phi}^* \mathbf{S}\bx(t) \right) = \boldsymbol{\Phi}^* \mathbf{S} \mathbf{y}(t)
    \label{eq:equivariance-a}
\end{equation}

\begin{figure}[ht]
\caption{Architecture of one layer of the FFT module, based on resnet, containing two convolutional layers with spectral normalization, activation functions (ReLU in our case), maxpooling with phase implementation, phase append module, and Fourier inverse module}
\centering
\includegraphics[width=0.8\textwidth]{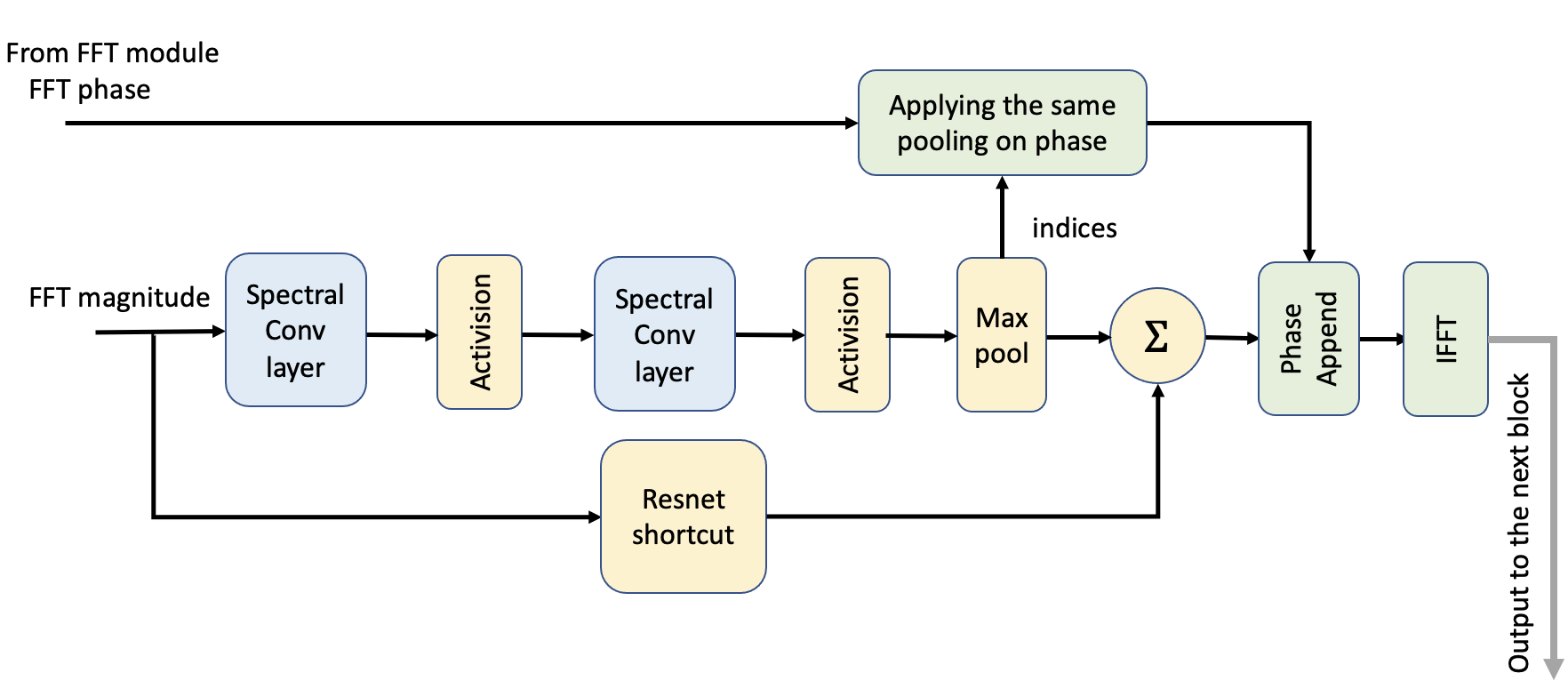}
\label{fig:fourierModule}
\end{figure}

In other words, the exact shift on input should be applied to the spatial domain image corresponding to the output Fourier spectrum. From figure \ref{fig:shiftEquivariance}-a we also have:
\begin{equation}
    \Tilde{f_f} \left( \boldsymbol{\Phi}^* \bx(t) \right) = \boldsymbol{\Phi}^* \mathbf{y}(t)
    \label{eq:equivariance-b}
\end{equation}
Fourier matrix $\boldsymbol{\Phi}^*$ is Unitary orthogonal hence, $\boldsymbol{\Phi \Phi}^* = \boldsymbol{\Phi}^* \boldsymbol{\Phi} = \mathbf{I}$, Thus from equation (\ref{eq:equivariance-b}) we get:
\begin{equation}
    \mathbf{y}(t) = \boldsymbol{\Phi} \Tilde{f_f} \left( \boldsymbol{\Phi}^* \bx(t) \right)
    \label{eq:output_rep}
\end{equation}
By putting equation (\ref{eq:output_rep}) in equation (\ref{eq:equivariance-a}):
\begin{equation}
     \Tilde{f_f} \left( \boldsymbol{\Phi}^* \mathbf{S}\bx(t) \right) =  \boldsymbol{\Phi}^* \mathbf{S} \boldsymbol{\Phi} \Tilde{f_f} \left( \boldsymbol{\Phi}^* \bx(t) \right)
     \label{eq:eqref-0}
\end{equation}
From properties of circulant matrices discussed in section \ref{sec:DFTGDL}, $\boldsymbol{\Phi}^*$ contains the eigenvectors of $\mathbf{S}$ in each row. Hence, $\boldsymbol{\Phi}^* \mathbf{S} = \boldsymbol{\Theta \Phi}^* $ whera $\boldsymbol{\Theta}$ is a diagonal matrix, consisting of $\mathbf{S}$'s eigenvalues on its diagonal. Knowing that we can rewrite equation (\ref{eq:eqref-0}) as:
\begin{equation*}
    \Tilde{f_f} \left( \boldsymbol{\Theta \Phi}^*\bx(t) \right) =  \boldsymbol{\Theta \Phi}^* \boldsymbol{\Phi} \Tilde{f_f} \left( \boldsymbol{\Phi}^* \bx(t) \right)
\end{equation*}
Finally we get to our desired constraint on the function $\Tilde{f_f}$:
\begin{equation}
    \Tilde{f_f}\left( \boldsymbol{\Theta} \fo\left( \bx(t) \right) \right) = \boldsymbol{\Theta} \Tilde{f_f}\left( \fo \left( \bx(t) \right) \right)
    \label{eq:final_constraint}
\end{equation}
$\mathbf{S}$ is non-symmetric orthogonal, hence its eigenvalues are complex and roots of $1$ with absolute value of $1$ and they are located and the diagonal of $\boldsymbol{\Theta}$. Thus, if we define $\Tilde{f_f}$ any function $\Tilde{f_f}: \mathbb{R}^n \to \mathbb{R}^n$ that keeps the phase intact, the equation (\ref{eq:final_constraint}) will be satisfied.\\
Now that we found a set of functions operating on the spectrum, keeping the spatial equivarincy, our next step is to design an architecture satisfying the resulted constraint. To that end, we first perform the Fourier transform and separate amplitude and phase. While we keep the phase intact, we pass the amplitude from a residual block with spectrum normalization. The spectral normalization is an attempt to stabilize GAN training introduced by \cite{sngan}. Performing max-pooling will change the dimension of the amplitude. We get the pooling indices and apply them to the phase to preserve the phase while sustaining its dimensions compatible with the amplitude. Since it is an element-wise operation over the phase matrix, it does not jeopardize the equivarincy. After all the computation on amplitude, we append the phase to get a complete spectrum representation. In the end, we perform a Fourier inverse to get back to the spatial domain and input to the next layer. Figure \ref{fig:fourierModule} shows the different elements of the architecture and their relationship together.

    \chapter{Experiments}
\label{chap:eval}
\section{Experimental Setup}
\label{sec:Setup}
\subsection{Systems and Baseline}
The systems used for the experiments are provided by the Laboratory for Neural Computing for Machine Learning, the specifications of the systems are listed in the table \ref{tab:systems}\\
\begin{table}[ht]
\centering
    \begin{tabular}{ |c||c|c|c|  }
        \hline
        \multicolumn{4}{|c|}{\textbf{System List}} \\
        \hline
        \textbf{System} & \textbf{CPU} & \textbf{Memory} & \textbf{GPUs}\\
        \hline
        Speech 8 & 4-core CPU (Xeon W-2125) & 32GB &   1 $\times$ GTX 1080, 8GB memory\\
        Text & 6-core CPU (i7-5820K) & 64GB & 4 $\times$ TITAN X, 12GB memory\\
        Video&6-core CPU (Xeon E5-1650)& 64GB&3 $\times$ TITAN X, 12GB memory\\
        Image&6-core CPU (i7-6800K)&128GB&4 $\times$ GTX 1080, 8GB memory\\
        \hline
    \end{tabular}
    \label{tab:systems}
    \caption{List of systems used for the experiments}
\end{table}
The experiments are all conducted with the python version 3.7, PyTorch version 1.7.1 utilized with CUDA 10.1, with torchvision version 0.8.2, and torchaudio version 0.7.2.\\
\textit{SNGAN} \cite{sngan} is the baseline chosen for the GAN experiments. It uses spectral normalization (normalization of the weight matrices by the largest singular value) in its convolutional layers. It is a fairly stable version of GANs, and it is used in other frequency-driven efforts for GAN improvements such as \cite{ssdgan}, so it makes our work comparable with others. The implementation is done based on \textit{Mimicry} \cite{mimicry} which is a lightweight PyTorch library aimed towards the reproducibility of GAN research. It is introduced in the \textit{CVPR} 2020 workshops to resolve (in their words): "(a) Standardized implementations of popular GANs that closely reproduce reported scores; (b) Baseline scores of GANs trained and evaluated under the same conditions; (c) A framework for researchers to focus on the implementation of GANs without rewriting most of GAN training boilerplate code, with support for multiple GAN evaluation metrics."\\
\subsection{Datasets}
We used two datasets in this work, CIFAR100 \cite{cifar10} and stl-10 \cite{stl-10}. The first one consists of 32$\times$32 images in 100 classes developed by the University of Toronto. It has 600 images in each class, which makes it a total of 60000 images. There are 500 training and 100 testing images in each class. Since in the unsupervised problem of GAN, we do not need a separate testing class, we used all 600 images for the training purpose. This dataset provides us with images with different features due to its variety in classes. It is important for us since our objective is to see the improvement of the GANs for the general problem of creating synthetic images, not as a class conditional problem. Some examples of the classes are animals, insects, fruits and vegetables, plants, household devices and furniture, people, plants, outdoor and indoor scenes with different lightings and vehicles. Stl-10 contains 10 classes of images: airplane, bird, car, cat, deer, dog, horse, monkey, ship, and truck. in each 500 training and 800 testing images which we used both for training. The resolution of the original images is 96$\times$96. However, we resized the images to 48$\times$48 for memory-related and comparison reasons. The images in this dataset are acquired from rescaling of the ImageNet \cite{imagenet} dataset.
\subsection{Optimization and GAN Hyperparameters Settings}
We used Adam optimizer for both generator and discriminator with the setting: $(\beta_0, \beta_1) = (0.0 , 0.9)$ . The learning rate is set to $0.0002$ with a linear decay throughout the training. The batch size is set to $64$ and we adopt Xavier initialization for the all neural network weights. We update discriminator 5 times for each generator update to keep the training balance. The architecure for generator for each dataset is reported in the table \ref{tab:cifarG_nn} and \ref{tab:stlG_nn}, for the discriminator spatial path in table \ref{tab:cifarDs_nn} and \ref{tab:stlDs_nn}, for the discriminator frequency path in table \ref{tab:cifarDf_nn} and \ref{tab:stlDf_nn} and finally for the last unitary module of the discriminator in tables \ref{tab:cifarDunit_nn} and \ref{tab:stlDunit_nn}. \\
The FFT input to the frequency path consists of a double-sided FFT of the gray-scaled images. We chose to make our images monochrome since color frequencies possess less importance than brightness frequencies. We needed our module to focus on the brightness frequencies for this step. The addition of the color frequencies to the path is in our plans for future works.\\
Notations for reading the tables are: ResBlock: A residual block with a compatible shortcut whether it has down or up a sample or not, CONV: a convolutional layer with the written specifications, SNCONV: a convolutional layer with spectral normalization, K: kernel size, S: stride size, P: padding size, N: number of output channels, FC: fully connected layer with number of input and output neurons, Up: an upsampling procedure, PhaseAttach: procedure of attaching the phase which is pooled the same as amplitude in parallel, BN: batch normalization
\begin{table}[H]
    \centering
    \begin{tabular}{l l}
    \hline
         \textbf{Layer}&\textbf{Input} $\to$ \textbf{Output Shape}  \\
         \hline
         FC-(128,2048), Reshape & (128) $\to$ (256,4,4)\\
         \hline
         ResBlock: SNCONV-(N256,K3,S1,P1), BN, ReLU, Up & (256,4,4) $\to$ (256,8,8) \\
         \hline
         ResBlock: SNCONV-(N256,K3,S1,P1), BN, ReLU & (256,8,8) $\to$ (256,8,8) \\
         \hline
         ResBlock: SNCONV-(N256,K3,S1,P1), BN, ReLU, Up & (256,8,8) $\to$ (256,16,16) \\
         \hline
         ResBlock: SNCONV-(N256,K3,S1,P1), BN, ReLU & (256,16,16) $\to$ (256,16,16) \\
         \hline
         ResBlock: SNCONV-(N256,K3,S1,P1), BN, ReLU, Up & (256,16,16) $\to$ (256,32,32) \\
         \hline
         ResBlock: SNCONV-(N256,K3,S1,P1), BN, ReLU & (256,32,32) $\to$ (256,32,32) \\
         \hline
         ResBlock: SNCONV-(N256,K3,S1,P1), BN, ReLU & (256,32,32) $\to$ (256,32,32) \\
         \hline
         ResBlock: SNCONV-(N256,K3,S1,P1), BN, ReLU & (256,32,32) $\to$ (256,32,32) \\
         \hline
         BN, ReLU, CONV-(N3,K3,S1,P1), Tanh & (256,32,32) $\to$ (3,32,32)\\
         \hline
    \end{tabular}
    \caption{Architecture for CIFAR-100 generator $\mathcal{G}$}
    \label{tab:cifarG_nn}
\end{table}

\begin{table}[H]
    \centering
    \begin{tabular}{l l}
    \hline
         \textbf{Layer}&\textbf{Input} $\to$ \textbf{Output Shape}  \\
         \hline
         FC-(128,18432), Reshape & (128)$\to$(512,6,6)\\
         \hline
         ResBlock: SNCONV-(N256,K3,S1,P1), BN, ReLU, Up & (512,6,6) $\to$ (256,12,12) \\
         \hline
         ResBlock: SNCONV-(N256,K3,S1,P1), BN, ReLU & (256,12,12) $\to$ (256,12,12) \\
         \hline
         ResBlock: SNCONV-(N128,K3,S1,P1), BN, ReLU, Up & (256,12,12) $\to$ (128,24,24) \\
         \hline
         ResBlock: SNCONV-(N128,K3,S1,P1), BN, ReLU & (128,24,24) $\to$ (128,24,24) \\
         \hline
         ResBlock: SNCONV-(N64,K3,S1,P1), BN, ReLU, Up & (128,24,24) $\to$ (64,48,48) \\
         \hline
         ResBlock: SNCONV-(N64,K3,S1,P1), BN, ReLU & (64,48,48) $\to$ (64,48,48) \\
         \hline
         ResBlock: SNCONV-(N64,K3,S1,P1), BN, ReLU & (64,48,48) $\to$ (64,48,48) \\
         \hline
         ResBlock: SNCONV-(N64,K3,S1,P1), BN, ReLU & (64,48,48) $\to$ (64,48,48) \\
         \hline
         BN, ReLU, CONV-(N3,K3,S1,P1), Tanh & (64,48,48) $\to$ (3,48,48)\\
         
    \end{tabular}
    \caption{Architecture for stl-10 generator $\mathcal{G}$}
    \label{tab:stlG_nn}
\end{table}

\begin{table}[H]
    \centering
    \begin{tabular}{p{10.5cm} p{4.75cm}}
    \hline
         \textbf{Layer}&\textbf{Input} $\to$ \textbf{Output Shape}  \\
        \hline
        ResBlock: SNCONV-(N128,K3,S1,P1), ReLU& (3,32,32)$\to$(128,32,32)\\
        \hline
        ResBlock: SNCONV-(N128,K3,S1,P1), ReLU, AvgPool-(K2,S2) & (128,32,32)$\to$(128,16,16)\\
        \hline
        ReLU, ResBlock: SNCONV-(N128,K3,S1,P1) & (128,16,16)$\to$(128,16,16)\\
        \hline
        ReLU, ResBlock: SNCONV-(N128,K3,S1,P1), AvgPool-(K2,S2) & (128,16,16)$\to$(128,8,8)\\
        \hline
        ReLU, ResBlock: SNCONV-(N128,K3,S1,P1) & (128,8,8)$\to$(128,8,8)\\
        \hline
        ReLU, ResBlock: SNCONV-(N128,K3,S1,P1) & (128,8,8)$\to$(128,8,8)\\
        \hline
        ReLU, ResBlock: SNCONV-(N128,K3,S1,P1) & (128,8,8)$\to$(128,8,8)\\
        \hline
        ReLU, ResBlock: SNCONV-(N128,K3,S1,P1) & (128,8,8)$\to$(128,8,8)\\
        \hline
        ReLU, GlobalSumPool & (128,8,8) $\to$ (128)
        
    \end{tabular}
    \caption{Architecture for CIFAR-100 Discriminator $\mathcal{D}$ spatial path}
    \label{tab:cifarDs_nn}
\end{table}

\begin{table}[H]
    \centering
    \begin{tabular}{p{10.5cm} p{4.75cm}}
    \hline
         \textbf{Layer}&\textbf{Input} $\to$ \textbf{Output Shape}  \\
        \hline
        ResBlock: SNCONV-(N64,K3,S1,P1), ReLU& (3,48,48)$\to$(64,48,48)\\
        \hline
        ResBlock: SNCONV-(N64,K3,S1,P1), ReLU, AvgPool-(K2,S2) & (64,48,48)$\to$(64,24,24)\\
        \hline
        ReLU, ResBlock: SNCONV-(N128,K3,S1,P1) & (64,24,24)$\to$(128,24,24)\\
        \hline
        ReLU, ResBlock: SNCONV-(N128,K3,S1,P1), AvgPool-(K2,S2) & (128,24,24)$\to$(128,12,12)\\
        \hline
        ReLU, ResBlock: SNCONV-(N256,K3,S1,P1) & (128,12,12)$\to$(256,12,12)\\
        \hline
        ReLU, ResBlock: SNCONV-(N256,K3,S1,P1), AvgPool-(K2,S2) & (256,12,12)$\to$(256,6,6)\\
        \hline
        ReLU, ResBlock: SNCONV-(N512,K3,S1,P1) & (256,6,6)$\to$(512,6,6)\\
        \hline
        ReLU, ResBlock: SNCONV-(N512,K3,S1,P1), AvgPool-(K2,S2) & (512,6,6)$\to$(512,3,3)\\
        \hline
        ReLU, ResBlock: SNCONV-(N1024,K3,S1,P1) & (512,3,3)$\to$(1024,3,3)\\
        \hline
        ReLU, ResBlock: SNCONV-(N1024,K3,S1,P1) & (1024,3,3)$\to$(1023,3,3)\\
        \hline
        ReLU, GlobalSumPool & (1024,3,3) $\to$ (1024)
    \end{tabular}
    \caption{Architecture for stl-10 Discriminator $\mathcal{D}$ spatial path}
    \label{tab:stlDs_nn}
\end{table}

\begin{table}[H]
    \centering
    \begin{tabular}{p{10.5cm} p{4.75cm}}
    \hline
         \textbf{Layer}&\textbf{Input} $\to$ \textbf{Output Shape}  \\
         \hline
         ResBlock: SNCONV-(N128,K3,S1,P1), ReLU& (1,32,32)$\to$(128,32,32)\\
         \hline
         ResBlock: SNCONV-(N128,K3,S1,P1), ReLU, MaxPool-(K2,S2)& (128,32,32)$\to$(128,16,16)\\
         \hline
         PhaseAttach, IFFT, GlobalSumPool & (128,16,16)$\to$(128)\\
         \hline
    \end{tabular}
    \caption{Architecture for CIFAR-100 Discriminator $\mathcal{D}$ frequency path}
    \label{tab:cifarDf_nn}
\end{table}

\begin{table}[H]
    \centering
    \begin{tabular}{p{10.5cm} p{4.75cm}}
    \hline
         \textbf{Layer}&\textbf{Input} $\to$ \textbf{Output Shape}  \\
         \hline
         ResBlock: SNCONV-(N256,K3,S1,P1), ReLU& (1,48,48)$\to$(256,48,48)\\
         \hline
         ResBlock: SNCONV-(N1024,K3,S1,P1), ReLU, MaxPool-(K2,S2)& (256,48,48)$\to$(1024,24,24)\\
         \hline
         PhaseAttach, IFFT, GlobalSumPool & (1024,24,24)$\to$(1024)\\
         \hline
    \end{tabular}
    \caption{Architecture for stl-10 Discriminator $\mathcal{D}$ frequency path}
    \label{tab:stlDf_nn}
\end{table}

\begin{table}[H]
    \centering
    \begin{tabular}{p{7cm} p{7cm}}
    \hline
         \textbf{Layer}&\textbf{Input} $\to$ \textbf{Output Shape}  \\
         \hline
         FC-(128,1) & (128) \textit{spatial} $\to$ (1) \textit{s-score} \\
         \hline
         FC-(128,1) & (128) \textit{frequency} $\to$ (1) \textit{f-score} \\
         \hline
         Concat & (128) \textit{spatial}, (128) \textit{frequency} $\to$ (256) \\
         \hline
         FC-(256,1) & (256) $\to$ (1) \textit{j-score}
    \end{tabular}
    \caption{Architecture for CIFAR-100 Discriminator $\mathcal{D}$ unitary modules}
    \label{tab:cifarDunit_nn}
\end{table}

\begin{table}[H]
    \centering
    \begin{tabular}{p{5cm} p{9cm}}
    \hline
         \textbf{Layer}&\textbf{Input} $\to$ \textbf{Output Shape}  \\
         \hline
         FC-(1024,1) & (1024) \textit{spatial} $\to$ (1) \textit{s-score} \\
         \hline
         FC-(1024,1) & (1024) \textit{frequency} $\to$ (1) \textit{f-score} \\
         \hline
         Concat & (1024) \textit{spatial}, (1024) \textit{frequency} $\to$ (2048) \\
         \hline
         FC-(2048,1) & (2048) $\to$ (1) \textit{j-score}
    \end{tabular}
    \caption{Architecture for stl-10 Discriminator $\mathcal{D}$ unitary modules}
    \label{tab:stlDunit_nn}
\end{table}

\section{Stability of The Training}
\label{sec:stability}
\begin{figure}[H]
\caption{Discriminator's $\mathcal{D}$ total loss and Generator's $\mathcal{G}$ loss learning curve during the 100000 iterations for CIFAR100 dataset}
\centering
\includegraphics[width=0.8\textwidth]{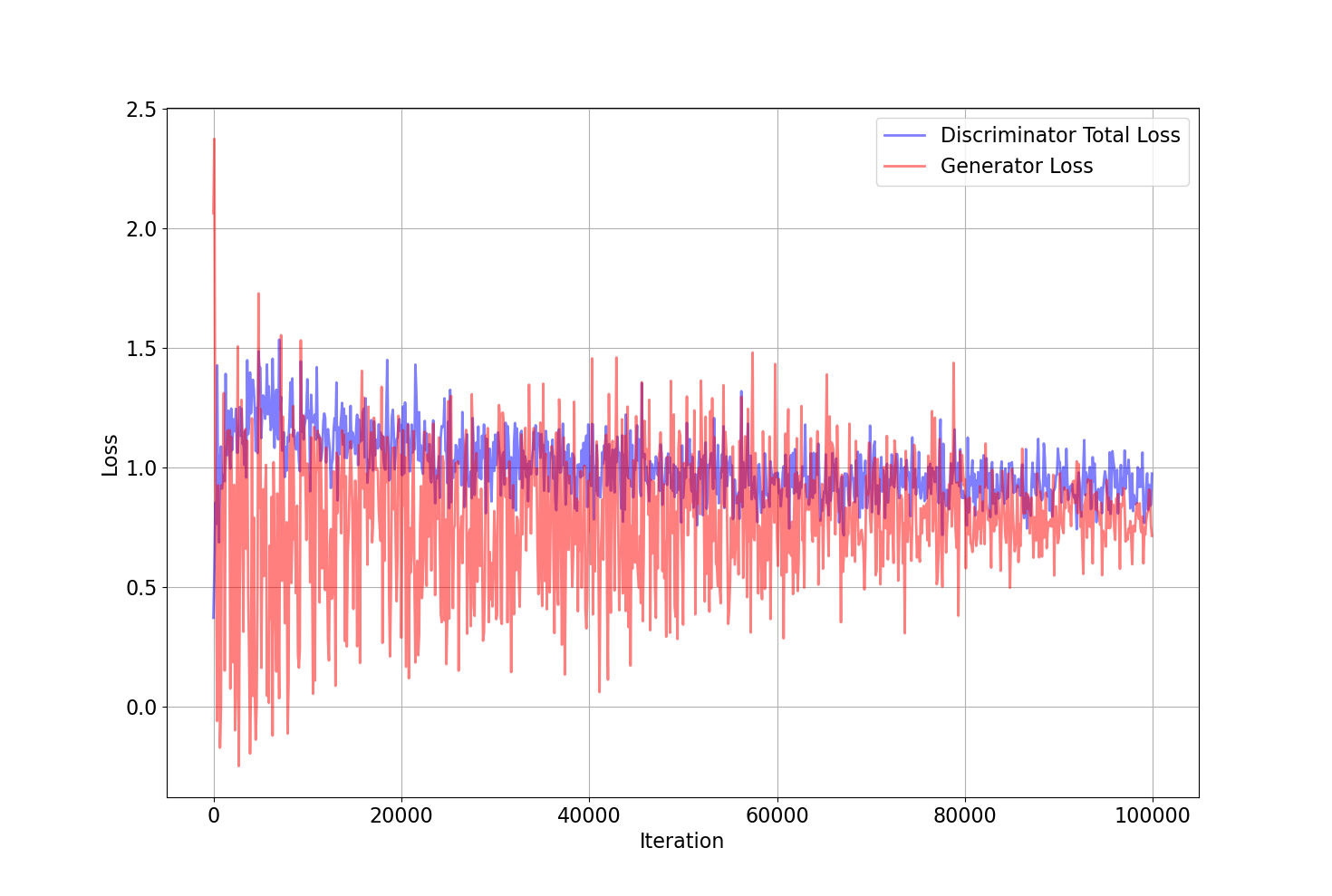}
\label{fig:cifar_DandG_loss}
\end{figure}

\begin{figure}[H]
\caption{Discriminator's $\mathcal{D}$ total loss and Generator's $\mathcal{G}$ loss learning curve during the 100000 iterations for stl10 dataset}
\centering
\includegraphics[width=0.8\textwidth]{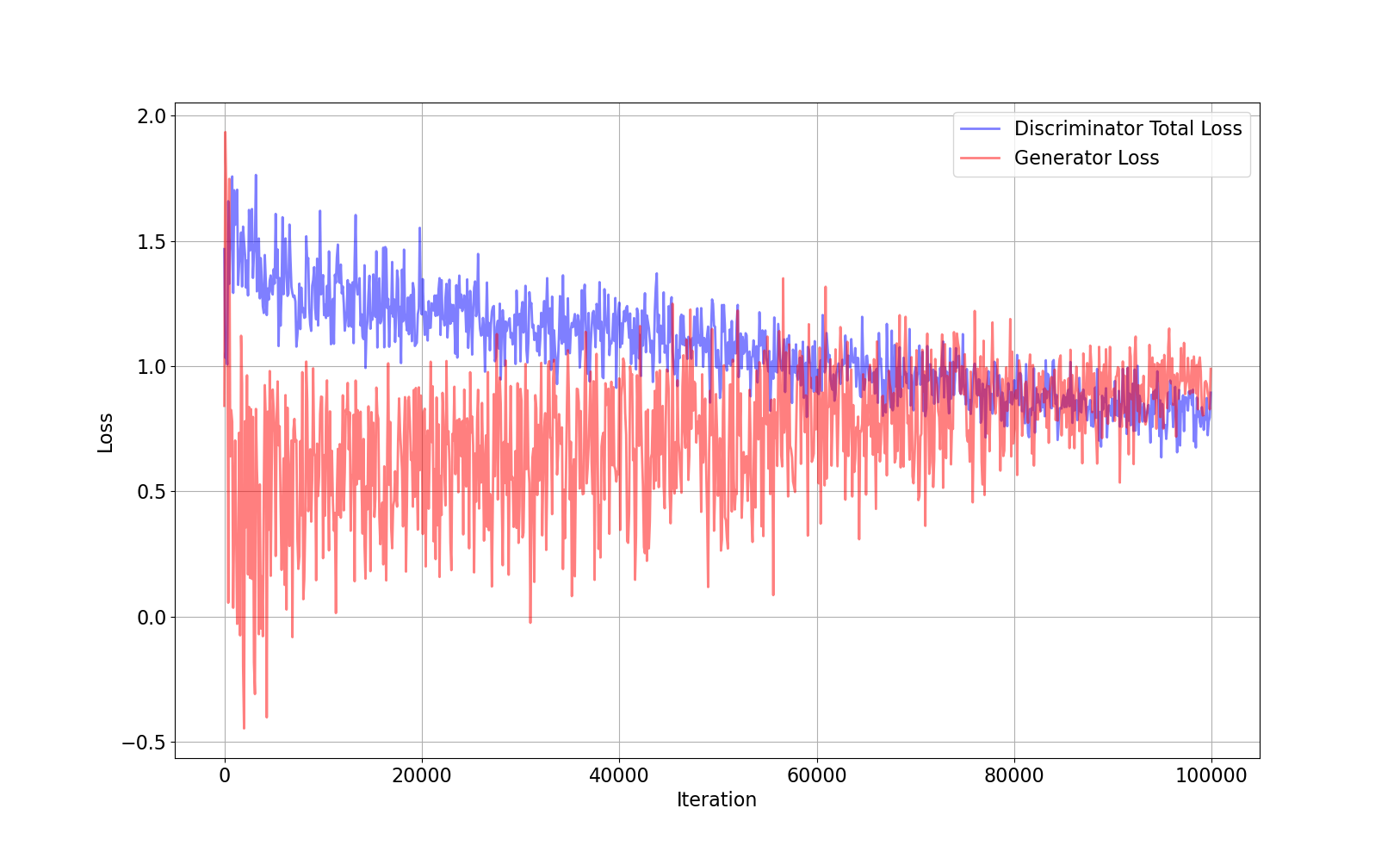}
\label{fig:stl_DandG_loss}
\end{figure}

\begin{figure}[H]
\caption{Detailed learning curves for loss of the 3 modules of the discriminator for CIFAR100 dataset.}
\centering
\includegraphics[width=0.8\textwidth]{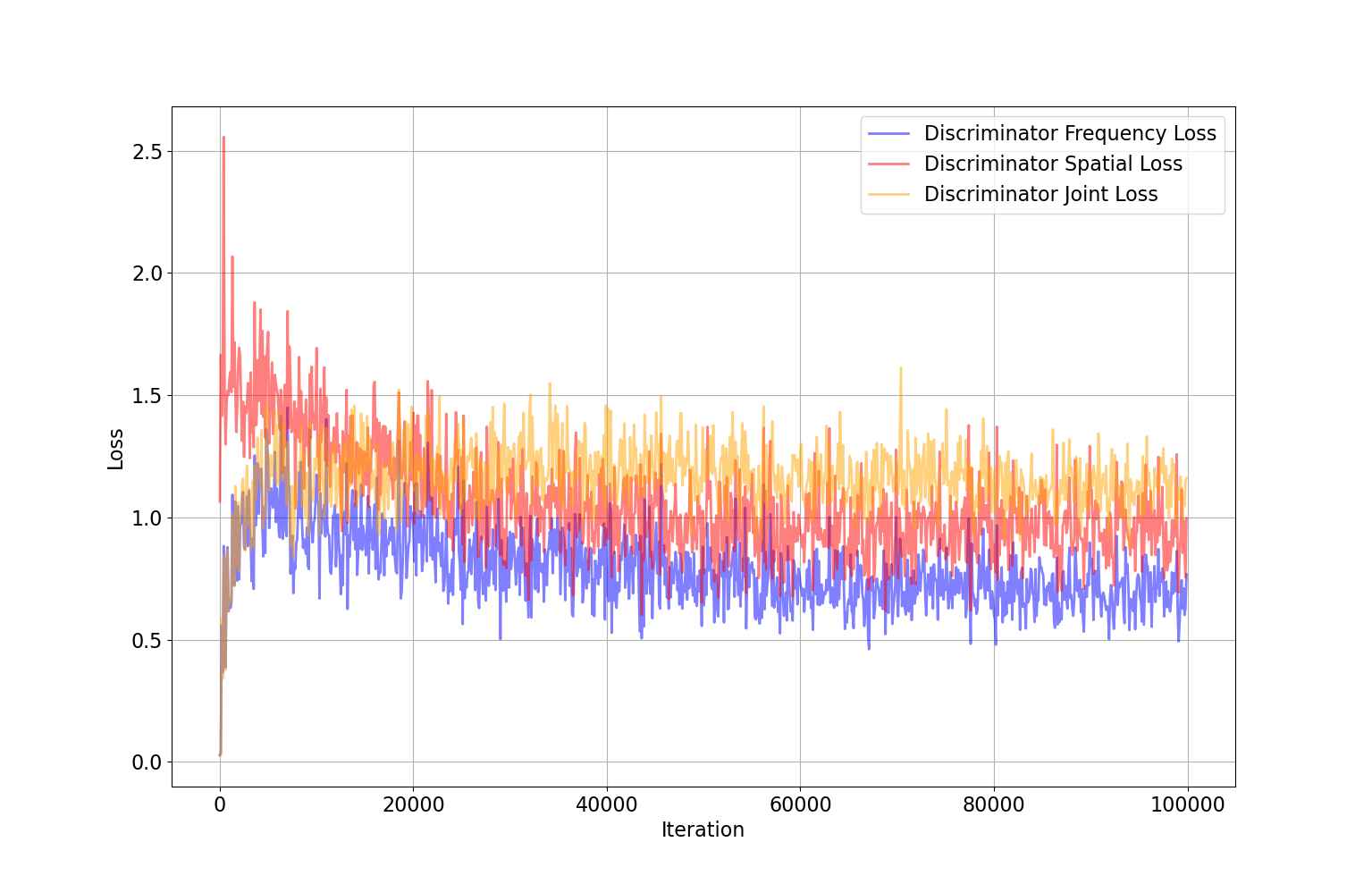}
\label{fig:cifar_module_loss}
\end{figure}

\begin{figure}[H]
\caption{Detailed learning curves for loss of the 3 modules of the discriminator for stl10 dataset.}
\centering
\includegraphics[width=0.8\textwidth]{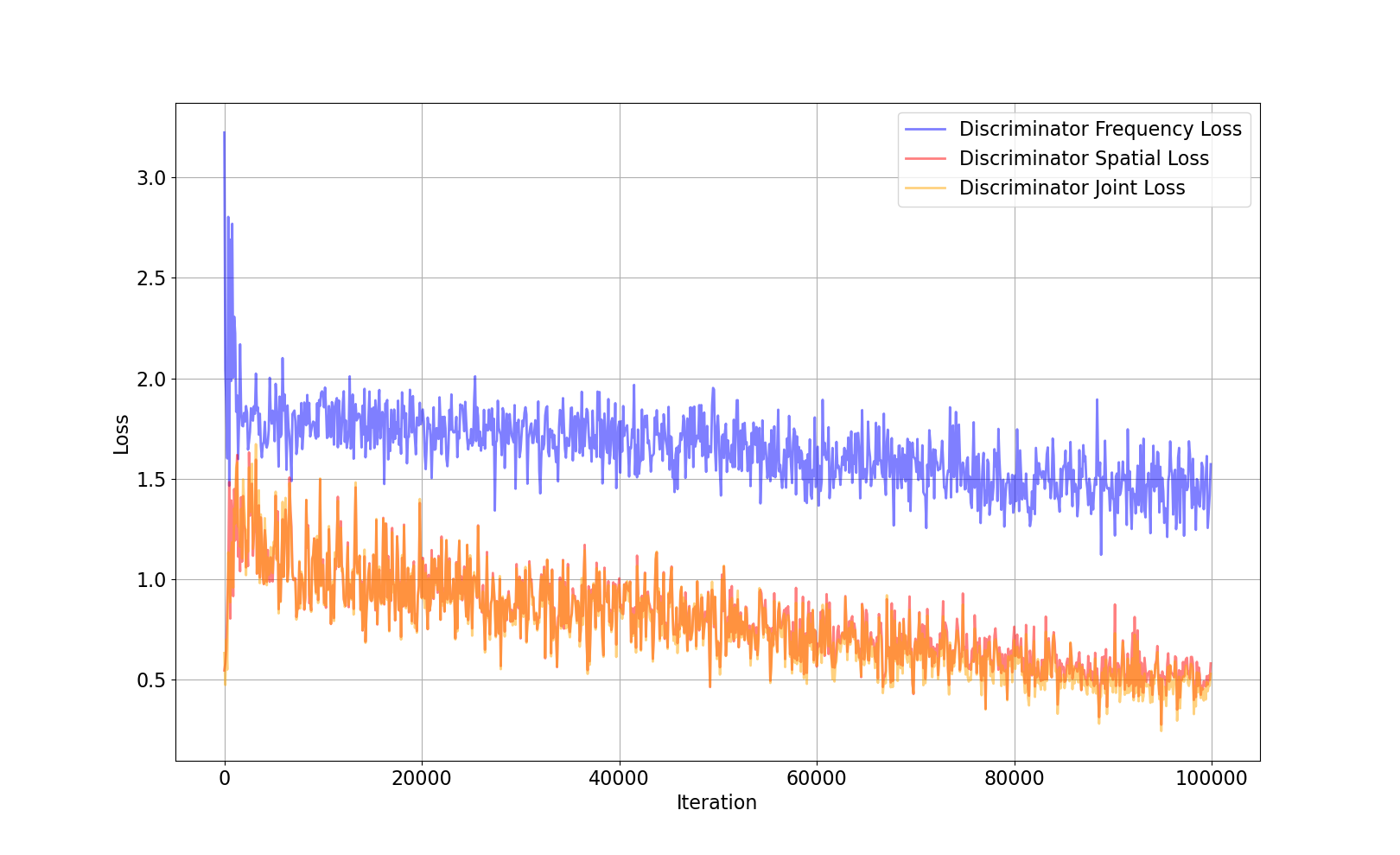}
\label{fig:stl_module_loss}
\end{figure}

\begin{figure}[H]
\caption{real and fake scores for each discriminator module during the training phase for CIFAR100 dataset.}
\centering
\includegraphics[width=0.8\textwidth]{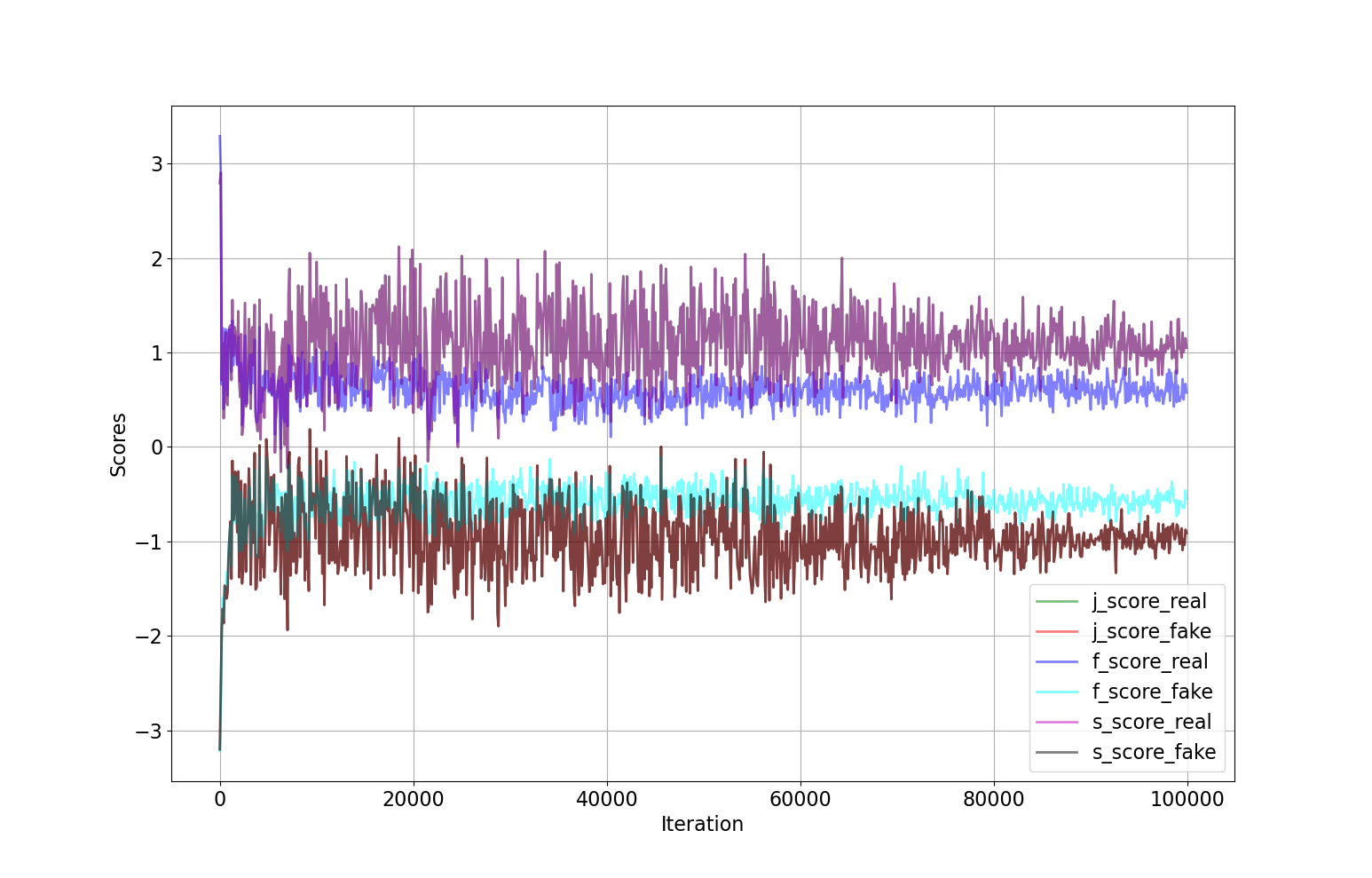}
\label{fig:cifar_scores}
\end{figure}

\begin{figure}[H]
\caption{real and fake scores for each discriminator module during the training phase for stl10 dataset.}
\centering
\includegraphics[width=0.8\textwidth]{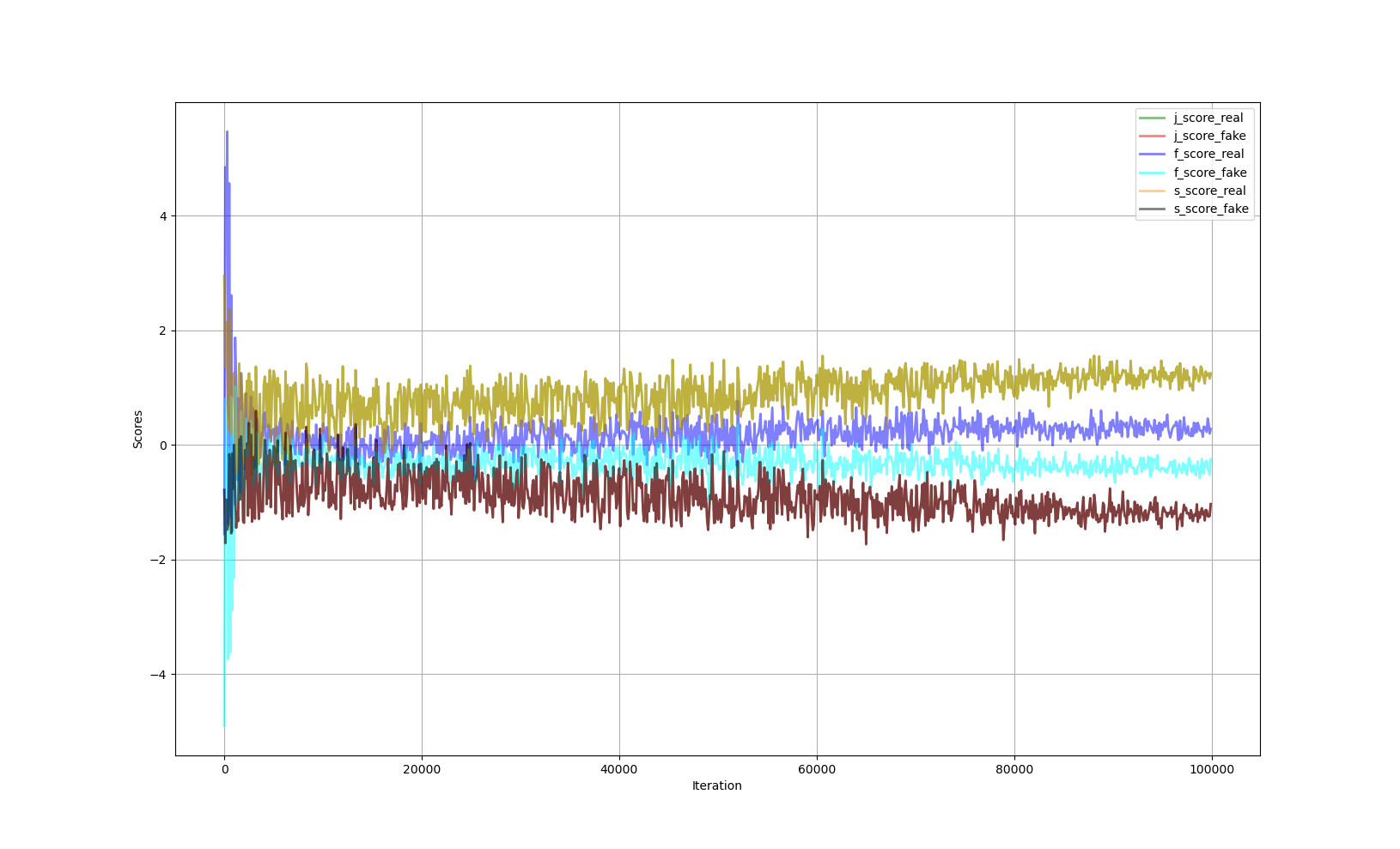}
\label{fig:stl_scores}
\end{figure}

Before evaluating any result, we need to ensure that the model converges. Otherwise, the results can not serve as an indication of the superiority of our model. In GAN tasks, unlike conventional tasks, the objective in training is not improving variables such as decreasing a loss. We care about a gradual and continuous improvement in both agents involved in the process, meaning generator and discriminator. If any agents get too strong with respect to the other, the balance is disturbed, and the model will fail. This multi-agent scheme made GANs more susceptible to the instability of training. In fact, GANs are famous for being hard to train due to instability, and there have been several works to make them more stable. In this section, we provide empirical substantiations that our extension to the GAN framework will not jeopardize the stability of the baseline. We further discuss how each of the spatial and frequency modules acts separately convergence-wise and how they contribute to the convergence of the discriminator in total.\\
When networks are out of balance, we see a fast convergence in one of them, showing that the converged network has won the competition. It can happen when one network is inherently more potent. To address this kind of problem, we update the weaker network more frequently. In our cases, the baseline updates the discriminator 5 times for each generator update, and our extension uses the same ratio without instability. The other cause of failure in GANs is mode collapse. A mode collapse refers to a generator model that can generate one or a small subset of different outcomes or modes. It happens when the generator finds a way to produce plausible outcomes for the discriminator without imitating the distribution. Mode collapse is easily detectable by looking at the generated images since they come from a small set of images and are not diverse. In the learning curve, it shows itself as an oscillatory behavior in the generator loss (and, in some cases, discriminator loss).\\
Figures \ref{fig:cifar_DandG_loss} and \ref{fig:stl_DandG_loss} show the Discriminator's total loss, sum of the discriminator's fake and real losses, and generator's loss together for CIFAR100 and stl10 datasets. The curves are in a balanced manner showing fair competition while converging to the same loss. There is no instability in the curves, whether a fast convergence or an oscillatory behavior. For a deeper look, each discriminator loss consists of 3 other losses coming from the spatial and frequency modules and one joint loss from the J module. Figures \ref{fig:cifar_module_loss} and \ref{fig:stl_module_loss} show the derailed losses for the mentioned datasets. Frequency loss has a lower value compared to the spatial loss in CIFAR100 and greater in the stl10. It shows that the frequency domain is as useful as the spatial domain to detect the fake images for the discriminator. The joint loss trains to gives most of its attention to the more functional loss. The power of unitary modules emerges here. While spatial and frequency losses equally contribute to the $\frac{1}{3}$ of the discriminator loss, the joint loss which is responsible for the last $\frac{1}{3}$ of the total loss, automatically learns to adopts its weights for better use of each domain.\\
Figures \ref{fig:cifar_scores} and \ref{fig:stl_scores} show the raw scores used by discriminator for classification. The real scores are expected to be above zero and fake scores below zero. Although joint score mostly follows the spatial scores, frequency scores show fewer fluctuations and well above and below the 0 threshold, making it a valid feature for classification.\\
To recapitulate, frequency features showed to be a good representative for fake and real images in the discriminator and made the discriminator more powerful. The Generator also shows the capacity to be pushed along with the frequency extension in the discriminator without losing stability.

\section{Evaluation in Spatial Domain}
\begin{figure}[H]
\caption{Random set of images generated based on cifar100 dataset}
\centering
\includegraphics[width=\textwidth]{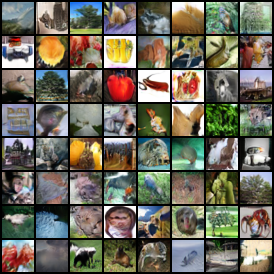}
\label{fig:sample_cifar}
\end{figure}

\begin{figure}[H]
\caption{Random set of images generated based on stl10 dataset}
\centering
\includegraphics[width=\textwidth]{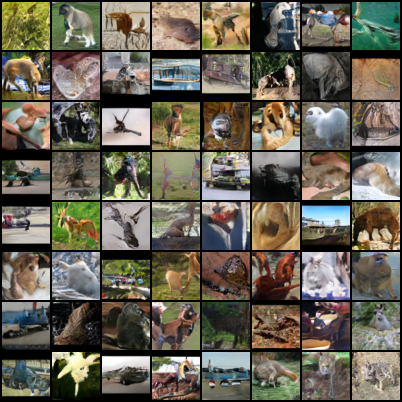}
\label{fig:sample_stl}
\end{figure}

We use the conventional methods, Frechet Inception Distance and Inception Score explained in the section \ref{sec:eval} for quantitative comparisons in the spatial domain. FID is calculated over 50000 samples of fake and real images. Inception score is computed over 50000 samples from the fake images with 10 splits.  The evaluation is against the baseline (\textit{SNGAN}) \cite{sngan} and the SSDGAN \cite{ssdgan} model. SSDGAN, the same as our model, uses frequency domain to enhance the discriminator. They use the azimuthal Integral of the amplitude as an input for their frequency module, which is a fully connected layer. Their model disregards the phase information of the FFT completely. While the amplitude of the FFT carries the intensity of each frequency, Phase information contains the location of these frequencies. To a human eyes phase information looks more important; hence we have not set them aside in our model. The other drawback of their model is using the azimuthal integration over FFT. By binning the same frequency resolutions, their model will lose the amplitude location information of the FFT. On the other hand, Ours takes the FFT as it is without any information loss as input.\\
table \ref{tab:CIFAR_compare} and \ref{tab:STL_compare} demonstrate the quantitative comparison. Our model outperforms all other models except the inception score for stl-10, which is less but comparable to SSDGAN. However, SSDGAN has some major flaws, which we fully describe in the next section. Note that for the FID, the lower, the better, and for inception score, the higher, the better.
\begin{table}[ht]
\centering
    \begin{tabular}{ |c||c|c|  }
        \hline
        \textbf{Model} & \textbf{FID} & \textbf{Inception Score}\\
        \hline
        SNGAN & 22.61 & 7.57 \\
        \hline
        SSDGAN & 20.90 & 7.71 \\
        \hline
        FreqGAN & \textbf{19.63} & \textbf{7.82} \\
        \hline
    \end{tabular}
    \label{tab:CIFAR_compare}
    \caption{FID and inception score comparison between different models for CIFAR-100 dataset}
\end{table}
\begin{table}[ht]
\centering
    \begin{tabular}{ |c||c|c|  }
        \hline
        \textbf{Model} & \textbf{FID} & \textbf{Inception Score}\\
        \hline
        SNGAN & 39.56 & 8.04 \\
        \hline
        SSDGAN & 36.41 & \textbf{8.47} \\
        \hline
        FreqGAN & \textbf{35.89} & 8.31 \\
        \hline
    \end{tabular}
    \label{tab:STL_compare}
    \caption{FID and inception score comparison between different models for stl-10 dataset}
\end{table}
Figure \ref{fig:sample_cifar} and \ref{fig:sample_stl} show some examples of the images created based on CIFAR100 and stl10 datasets respectively. We have not cherry picked the images and they are random images generated by the generator of our model.
\section{Evaluation in Frequency Domain}
\label{sec:FreqEval}
\begin{figure}[H]
\caption{Power spectrum of the real images, FreqGAN, SNGAN and SSDGAN with respect to the frequency bin in different resolutions on the stl10 dataset.}
\centering
\includegraphics[width=0.8\textwidth]{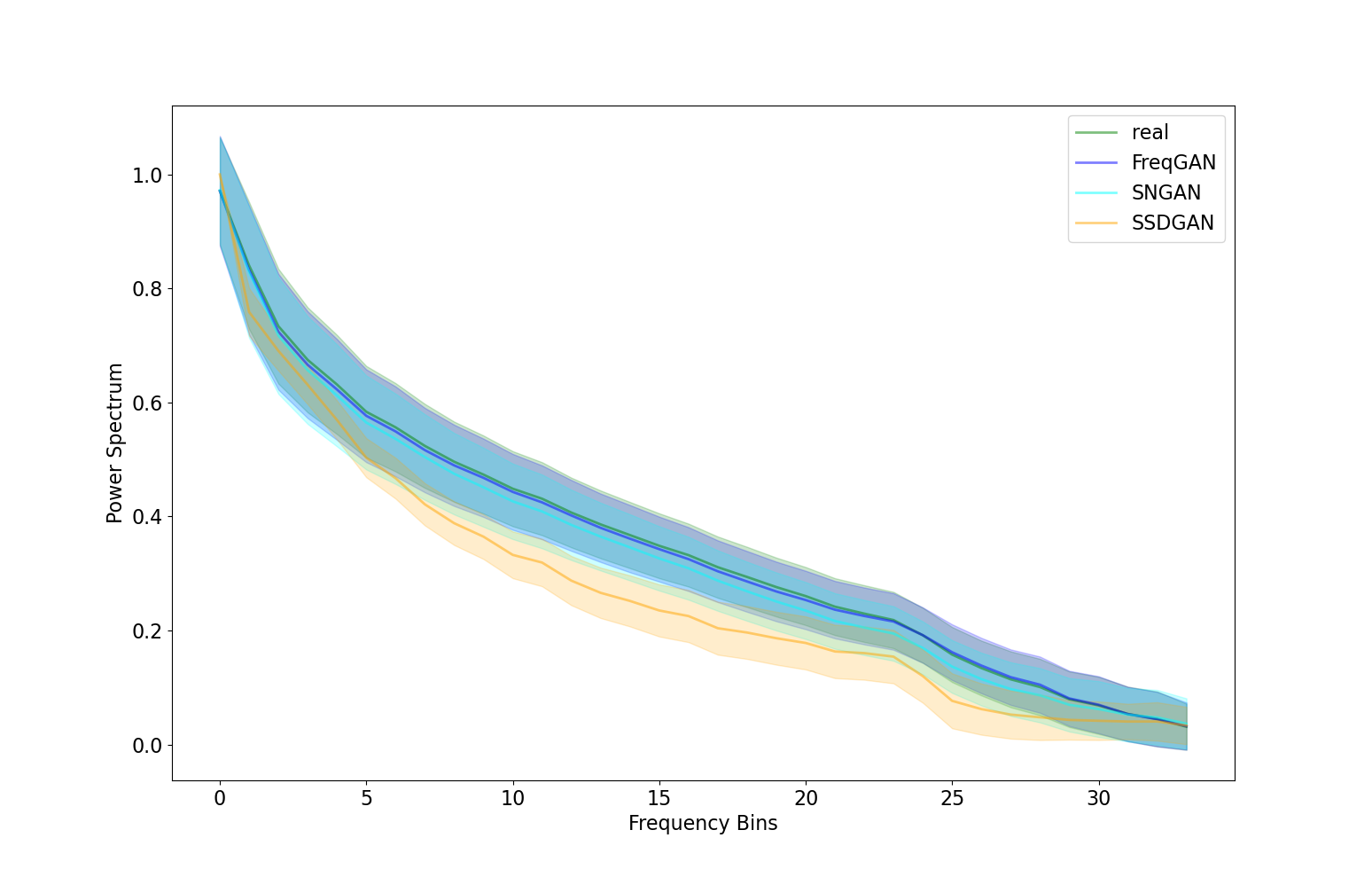}
\label{fig:stl10_power_spectrum}
\end{figure}

\begin{figure}[H]
\caption{Normalized power spectrum distance of FreqGAN, SNGAN and SSDGAN with respect to the frequency bin in different resolutions on the stl10 dataset. }
\centering
\includegraphics[width=0.8\textwidth]{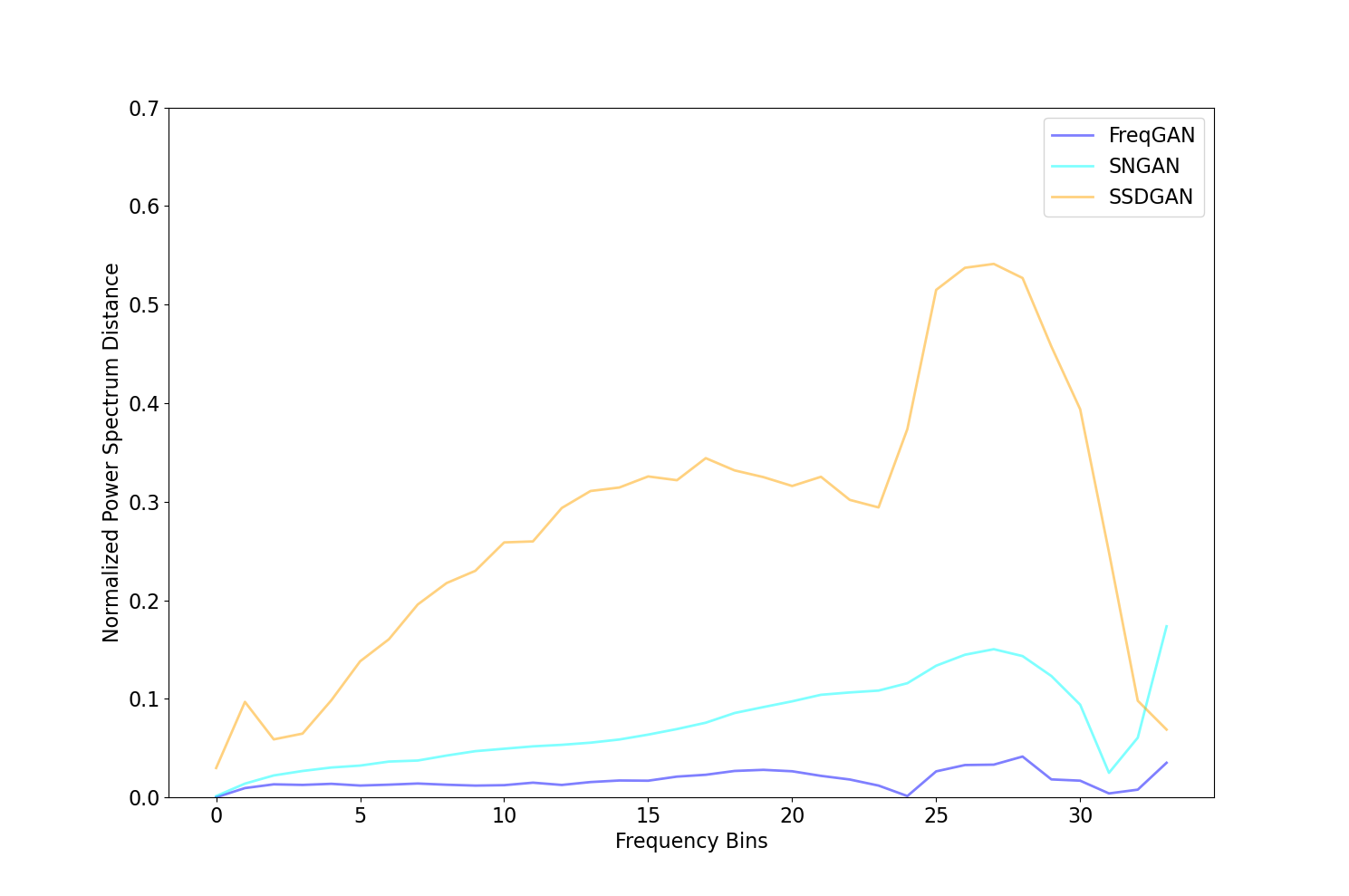}
\label{fig:stl10_power_spectrum_normalized}
\end{figure}

\begin{figure}[H]
\caption{2D frequency amplitude gap of FreqGAN, SNGAN and SSDGAN with respect to the frequency bin in different resolutions on the stl10 dataset. }
\centering
\includegraphics[width=\textwidth]{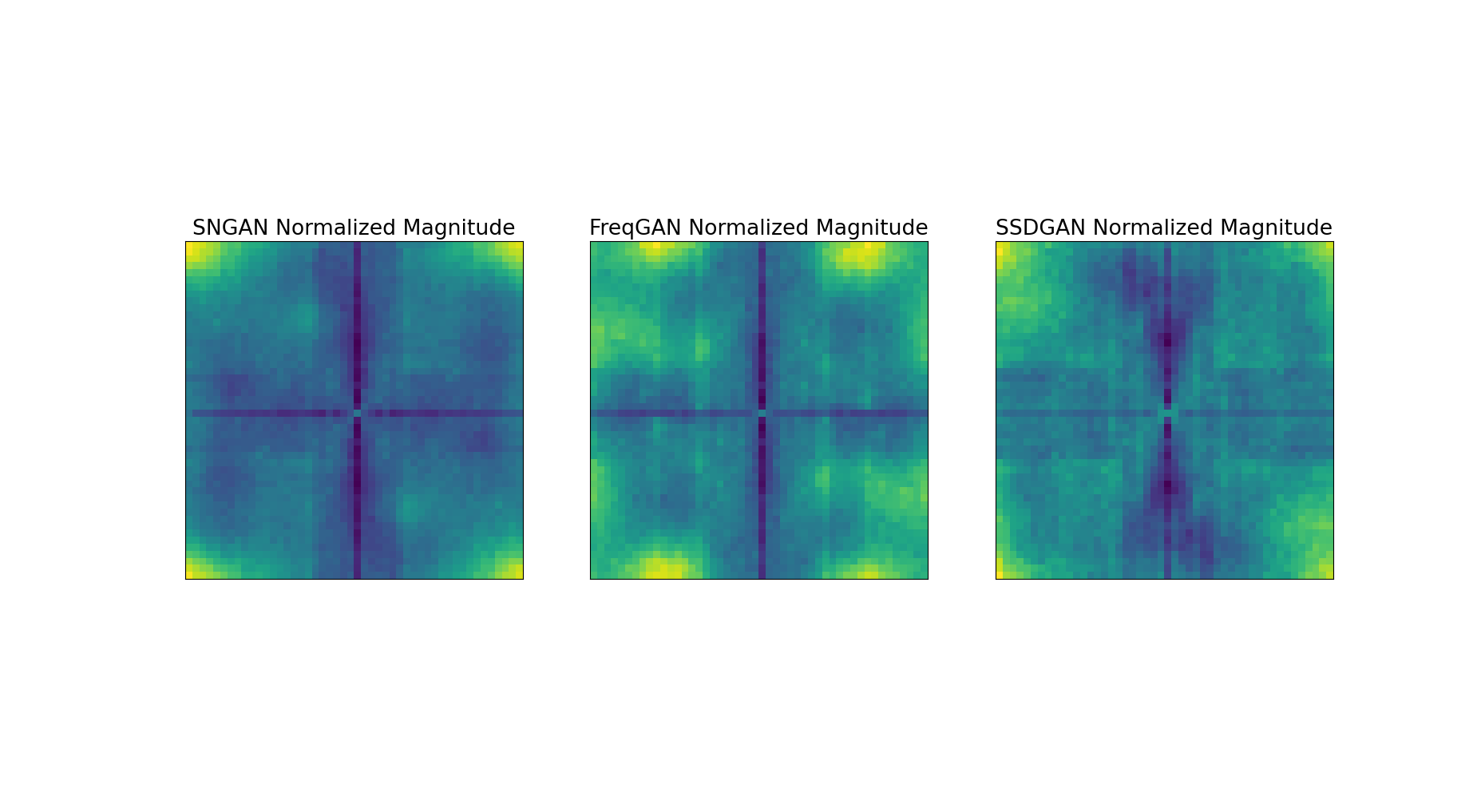}
\label{fig:stl10_spectrum_normalized}
\end{figure}

Unlike the spatial domain, Evaluation in the frequency domain does not have a well-established method. To illustrate how generated images are similar to the real ones in the frequency-domain power spectrum is a good candidate. To further show the difference, we use the power spectrum distance between real images and those produced by the GAN models, normalized by the spectra of real images. (used in \cite{swagan})
\begin{equation*}
    Power \: Spectrum \: Distance = \frac{|\mathbb{E}[PS(real))]-\mathbb{E}[PS(fake)]|}{\mathbb{E}[PS(real))]}
\end{equation*}
We also introduce the 2D frequency amplitude gap, which uses the average distance between the expected 2D-frequency gap between the amplitude of the authentic images and synthetic images normalized by those of authentic images.
\begin{equation*}
    2D\: Frequency\: Amplitude\: Gap = \frac{|\mathbb{E}[Amp(\fo(real))]-\mathbb{E}[Amp(\fo(fake)]|}{\mathbb{E}[Amp(\fo(real))]}
\end{equation*}
Figure \ref{fig:stl10_power_spectrum} shows the power spectrum of our model with real images and the other two models in the stl10 dataset. FreqGAN has the closest distance from the real image. It is more clear looking at the normalized power spectrum distance in figure \ref{fig:stl10_power_spectrum_normalized}. The ideal generator should have a constant zero distance throughout all the frequency bins. To our surprise, despite getting the azimuthal integration of the magnitude, which is essentially a variation of the power spectrum itself, SSDGAN has even poorer performance than the baseline SNGAN in the power spectrum. We believe this is owing to fully connected architecture. The fully connected neural networks have a dense hypothesis class in the frequency domain; therefore, estimating the best function even if the neurons are not many is hard for the neural network. In addition, the lack of unitary modules will cause an uncontrolled effect of these not-well-trained elements in the training process. FreqGAN, contrastingly, shows an almost flat curve over all frequency bins. addressing the high-frequency discrepancies in the SNGAN.\\
Figure \ref{fig:stl10_spectrum_normalized} measures the same idea in a more direct way with the normalized 2d frequency spectrum gap. The color map shows the difference; the darker the pixel, the more significant the difference between the fake and authentic images. FreqGAN here shows the best performance overall again, while SSDGAN shows inferior performance than the baseline.\\
For further discussion, we believe the difference of our work will show itself for higher resolution photos more significantly. The high-resolution photos suffer more from high-frequency discrepancies, and our model shows a spectacular result in addressing them. In our plans for future work, we have in mind to further analyze the relation of resolution with the performance of our model.

\section{Frequency Architecture Investigation}
\label{sec:GDL_arch}
We used the powerful tool of geometric deep learning to design our architecture for the frequency domain. In this section, we discuss its performance over the fully connected architecture and regular CNN. We trained 3 trails of neural networks for each architecture. The fully connected version is the same as the work done by SSDGAN \cite{ssdgan}. The CNN version is equipped with spectral normalization as well to keep the comparison fair. Tabel \ref{tab:trial_compare} shows that GDL based architecture outperforms other models by a significant margin. 

\begin{table}[ht]
\centering
    \begin{tabular}{ |c||c|  }
        \hline
        \textbf{Model} & \textbf{FID}\\
        \hline
        SNGAN (Baseline) & 22.61 \\
        \hline
        Fully Connected & 21.57 \\
        \hline
        Unmodified CNN & 22.01 \\
        \hline
        GDL inspired CNN & \textbf{19.63} \\
        \hline
    \end{tabular}
    \label{tab:trial_compare}
    \caption{FID comparison between different architectures in the frequency domain for CIFAR100 dataset}
\end{table}

    \chapter{Conclusion}
\label{chap:conclusion}

\section{Conclusion}
This thesis brought up current flaws in the CNN-based image generation techniques (such as generative adversarial networks), which made them easily distinguishable from authentic images. Then we investigate the possible justification for such flaws both mathematically and empirically. We found the main reason in high-frequency discrepancies between the spectrum of synthetic and authentic images. We showed that this disparity has its roots in the transconvolution layers of CNNs. This work proposed a solution for the GANs. An additional path for the discriminator, processing the spectrum of the images. Knowing the frequency representation of the images, the discriminator is equipped with direct access to detect problematic high-frequency elements. On the other side of the story, the generator adopts its weight to reduce the high-frequency elements, resulting in more realistic, hard to distinguish images.\\
Different datasets exhibit different distributions in their spatial and frequency features. Some are easier to detect in frequency, some in spatial. The proposed unitary modules in this work adjust the amount of attention the model gives to frequency or spatial domains based on their importance in the data. This scheme in leading the optimization not only resulted in better outputs but also stabilized the training.\\
This work empirically investigates different architectural designs for the frequency path of the discriminator, including fully connected and CNN schemes, and comes up with a brand new architecture EV-Freq. EV-Freq has systematically designed based on the geometric deep learning blueprint. It utilizes the physical constraints of shift invariancy in the images as an inductive bias to reduce the search space and improve the neural network's estimation ability on the spectrum of the corresponding images. This thesis provides complete theoretical analysis and proofs for the newly design architecture.\\
We evaluate our extension to the GANs framework over two datasets, CIFAR100 and stl10, in spatial and frequency domains. In the conventional spatial domain, it outperformed the baseline by a significant margin. Our model also outperformed the other frequency-driven effort in improving GANs, SSDGAN. FreqGAN achieved an outstanding performance in the frequency domain, generating images almost identical to real images concerning the power spectrum. GANs are notorious for training instability. A section of this work focused on empirical substantiations that the proposed extension would not jeopardize the stability of the baseline.

\section{Future Works}

The future steps for the current research are listed as follow:
\begin{itemize}
  \item This work improves the quality spectrum characteristics of the latent distribution of the generator by improving the discriminator power in frequency. Yet, it does not propose any direct change in the generator's architecture. Investigating new designs for the generator to facilitate adopting with the frequency features is one of our prominent future lines of research 
  \item EV-freq introduced a shift equivariant function operating on frequency domain; however, these classes of functions keep the phase information intact, in the future steps, we continue our search to find functions operating on the phase of FFT spectrum to utilize them in the detection process of the discriminator.
  \item Current work uses the FFT of the grayscaled version of the images. In other words, it just takes advantage of the brightness frequency in the images. For future steps, we intend to extend our work to a colored version of the image utilizing the color frequency information in our model
  \item New PyTorch update introduces complex autograd. Since back-propagation is now available for complex numbers, we can readily work with complex losses. It promises a new architectural design with no need for IFFT at the end of the block, which can improve our performance in the spectral domain and improve the memory usage and computation complexity.
\end{itemize}
    
    \printbibliography [heading=bibintoc, title={Bibliography}]
    % \bibliographystyle{acm.bst}
    % \bibliography{Bibliography}
    % \printbibliography
    % main text follows
    %% include your chapters here. 
    %% include your appendix here.
\end{document}